\documentclass[journal]{IEEEtran}
\usepackage{amsmath}
\usepackage{cite}
\usepackage{graphicx}
\usepackage{epstopdf} 
\usepackage{caption}
\usepackage{amsfonts}
\usepackage{quotes}
\usepackage{subcaption}
\usepackage{subfig}
\usepackage{fancyhdr}
\usepackage{amssymb}
\usepackage{color}
\usepackage{algorithmic}

\ifCLASSINFOpdf
\else
\fi

\newcommand{\Ls}{{\mathbf{L}\boldsymbol{_s}}}
\newcommand{\Uivec}{{\mathbf{u}_i}}

\begin{document}

\title{Multiclass Data Segmentation using Diffuse Interface Methods on Graphs}

\author{Cristina~Garcia-Cardona,
        Ekaterina~Merkurjev,
        Andrea~L.~Bertozzi, 
        Arjuna~Flenner and Allon~G.~Percus

\thanks{C.~Garcia-Cardona and A.G.~Percus are with the Institute of
Mathematical Sciences at Claremont Graduate University. E-mail:
cristina.cgarcia@gmail.com, allon.percus@cgu.edu.}
\thanks{E.~Merkurjev and A.L.~Bertozzi are with the Department of Mathematics at University of California, Los Angeles. Email: \{kmerkurev,bertozzi\}@math.ucla.edu.}
\thanks{A.~Flenner is with the Naval Air Warfare Center, China Lake.}
}

\IEEEaftertitletext{\vspace{-1\baselineskip}
\noindent
\begin{abstract}

\indent \indent 
We present two graph-based algorithms for multiclass segmentation of
high-dimensional data on graphs.
The algorithms use a diffuse interface model based on the
Ginzburg-Landau functional, related to total variation and graph cuts.
A multiclass
extension is introduced using the Gibbs simplex, with the functional's
double-well potential modified
to handle the multiclass case. The first algorithm minimizes the functional
using a convex splitting numerical scheme. The second algorithm 
uses a graph adaptation of the classical numerical
Merriman-Bence-Osher (MBO) scheme, which alternates between diffusion
and thresholding.
We demonstrate the performance of both algorithms experimentally on synthetic data, image labeling, and several benchmark
data sets such as MNIST, COIL and WebKB. We also make use of fast numerical solvers for finding the eigenvectors and eigenvalues 
of the graph Laplacian, and take advantage of the sparsity of the matrix. Experiments indicate that the results are
 competitive with or better than the current state-of-the-art in
multiclass graph-based segmentation algorithms for high-dimensional
data.

\end{abstract}
\noindent

\vspace{1\baselineskip}}

\maketitle

\IEEEpeerreviewmaketitle

\section{Introduction}

Multiclass segmentation is a fundamental problem in 
machine learning.  In this paper, we present a general approach to multiclass
segmentation of high-dimensional data on graphs, motivated by the
diffuse interface model in~\cite{bertozzi:flenner}. The method
applies $L_2$ gradient flow minimization of the Ginzburg-Landau (GL)
energy to the case of functions defined on graphs.

The GL energy is a smooth functional that converges, in the limit of a
vanishing interface width, to the total variation
(TV)~\cite{kohn,bertozzi:vangennip}.  There is a close connection
between TV minimization and graph cut minimization.  Given
a graph $G=(V,E)$ with vertex set $V$, edge set $E$, and edge weights
$w_{ij}$ for $i,j \in V$, the TV norm of a function
$f$ on $V$ is
\begin{equation}
	|| f ||_{TV} = \frac{1}{2} \sum_{i,j \in V} w_{ij} | f_i - f_j |.
\end{equation}
If $f_i$ is interpreted as a classification of vertex $i$, minimizing TV is
exactly equivalent to minimizing the graph cut.  TV-based methods have
recently been used~\cite{szlam,bresson:laurent,bresson:laurent:2013} to
find good approximations for normalized graph cut minimization, an
NP-hard problem.  Unlike methods such as
spectral clustering, normalized TV minimization provides a tight
relaxation of the problem, though cannot usually be solved
exactly.  The approach in~\cite{bertozzi:flenner} performs binary
segmentation on graphs by using the GL functional as a smooth but
arbitarily close approximation to the TV norm.




Our new formulation builds on~\cite{bertozzi:flenner}, using a
semi-supervised learning (SSL) framework for multiclass graph segmentation.
We employ a phase-field representation of the GL energy
functional: a vector-valued quantity is assigned to every node on the
graph, such that each of its components represents the fraction of the phase,
or class, present in that particular node. The components of the field
variable add up to one, so the phase-field vector is constrained to lie
on the Gibbs simplex. The phase-field representation, used in 
material science to study the evolution of multi-phase
systems~\cite{garcke}, has been studied previously for multiclass image
segmentation~\cite{Lellmann2008}.  Likewise, the simplex idea has been
used for image segmentation~\cite{chambolle:cremers:pock,grady06}.
However, to the best of our knowledge, our diffuse interface
approach is the first application of a
vector-field GL representation to the general problem
of multiclass semi-supervised classification of high-dimensional data on
graphs.

In addition,
we apply this Gibbs simplex idea to the graph-based Merriman-Bence-Osher
(MBO) scheme developed in~\cite{merkurjev}.  The MBO scheme~\cite{MBO}
is a well-established PDE method for evolving an interface by mean
curvature. As with the diffuse interface model, tools for nonlocal
calculus~\cite{gilboa} are used in~\cite{merkurjev} to generalize the
PDE formulation to the graph setting. By introducing the phase-field
representation to the graph-based MBO scheme, we develop another new and
highly efficient algorithm for multiclass segmentation in a SSL framework. 

The main contributions of our work are therefore twofold.  First, we
introduce two new
graph-based methods for multiclass data segmentation, namely a
multiclass GL minimization method based on the binary representation described
in~\cite{bertozzi:flenner} and a multiclass graph-based MBO method
motivated by
the model in~\cite{merkurjev}.  Second, we present very efficient
algorithms derived from these methods, and applicable to general
multiclass high-dimensional data segmentation.


The paper is organized as follows. In section \ref{background}, we
discuss prior related work, as well as motivation for the methods proposed here.
We then describe our two new multiclass algorithms
in section \ref{first_algorithm} (one in section III-A and one in III-B). In section
\ref{results}, we present experimental results on benchmark data sets,
demonstrating the effectiveness of our
methods. Finally, in section \ref{conclusions}, we conclude and discuss
ideas for future work.



\section{Previous Work}
\label{background}

\subsection{General Background}

 In this section, we present prior related work, as well as specific algorithms that serve as motivation for our new multiclass methods. 

The discrete graph formulation of GL energy minimization is an example of a more general form of energy (or cost) functional for data classification in machine learning,
\begin{eqnarray}
E(\psi) =  R(\psi) + \mu || \psi - \hat{\psi} ||,
\label{eqn:basic}
\end{eqnarray}
where $\psi$ is the classification function, $R(\psi)$ is a regularization term, and $||\psi - \hat{\psi}||$ is a fidelity term, incorporating most (supervised) or just a few (semi-supervised) of the known values $\hat{\psi}$. The choice of $R$ has non-trivial consequences in the final classification accuracy. In instances where $|| \cdot ||$ is the $L_2$ norm, the resulting cost functional is a tradeoff between accuracy in the classification of given labels and function smoothness. It is desirable to choose $R$ to preserve the sharp discontinuities that may arise in the boundaries between classes. Hence the interest in formulations that can produce piecewise constant solutions~\cite{boykov:veksler}. 

Graph-based regularization terms, expressed by way of the discrete Laplace operator on graphs, are often used in semi-supervised learning as a way to exploit underlying similarities in the data set~\cite{zhou:bousquet:lal,zhou:scholkopf,wang,belkin,chapelle:scholkopf:zien,zhu}. Additionally, some of these methods use a matrix representation to apply eq. (\ref{eqn:basic}) to the multiple-class case~\cite{chapelle:scholkopf:zien,wang,zhu,zhou:bousquet:lal}. The rows in the matrix correspond to graph vertices and the columns to indicator functions for class membership: the class membership for vertex $i$ is computed as the column with largest component in the $i$th row. The resulting minimization procedure is akin to multiple relaxed binary classifications running in parallel. This representation is different from the Gibbs simplex we use, as there is usually no requirement that the elements in the row add up to 1. An  alternative regularization method for the graph-based multiclass setup is presented in~\cite{subramanya}, where the authors minimize a Kullback-Leibler divergence function between discrete probability measures that translates into class membership probabilities.

Not all the methods deal directly with the multiple classes in the data set. A different approach is to reduce the multiclass case to a series of two-class problems and to combine the sequence of resulting sub-classifications. Strategies employed include recursive partitioning, hierarchical classification and binary encodings, among others. For example, Dietterich and Bakiri use a binary approach to encode the class labels~\cite{dietterich}. In~\cite{hastie}, a pairwise coupling is described, in which each two-class problem is solved and then a class decision is made combining the decisions of all the subproblems. Szlam and Bresson present a method involving Cheeger cuts and split Bregman iteration \cite{goldstein} to build a recursive partitioning scheme in which the data set is repeatedly divided until the desired number of classes is reached. The latter scheme has been extended to mutliclass versions. In~\cite{bresson:tai}, a multiclass algorithm for the transductive learning problem in high-dimensional data classification, based on $\ell^1$ relaxation of the Cheeger cut and the piecewise constant Mumford-Shah or Potts models, is described. Recently, a new TV-based method for multiclass clustering has been introduced in~\cite{bresson:laurent:2013}.

Our methods, on the other hand, have roots in the continuous setting as
they are derived via a variational formulation. Our first method comes
from a variational formulation of the $L_2$ gradient flow minimization
of the GL functional~\cite{bertozzi:flenner}, but which in a limit
turns into $TV$ minimization.
Our second method is built
upon the MBO classical scheme to evolve interfaces by mean
curvature~\cite{MBO}. The latter has connections with the work presented
in~\cite{esedoglu}, where an MBO-like scheme is used for image
segmentation. The method is motivated by the propagation of the
Allen-Cahn equation with a forcing term, obtained by applying gradient
descent to minimize the GL functional with a fidelity term. 

Alternative
variational principles have also been used for image segmentation.
In~\cite{Lellmann2008}, a multiclass labeling for image analysis is
carried out by a multidimensional total variation formulation involving
a simplex-constrained convex optimization. In that work, a
discretization of the resulting PDEs is used to solve numerically the minimization of the energy. Also, in~\cite{chambolle:cremers:pock} a partition of a continuous open domain in subsets with minimal perimeter is analyzed. A convex relaxation procedure is proposed and applied to image segmentation. In these cases, the discretization corresponds to a uniform grid embedded in the Euclidean space where the domain resides. Similarly, diffuse interface methods have been used successfully in image impainting~\cite{bertozzi:esedoglu:gillette, dobrosotskaya:bertozzi:08} and image segmentation~\cite{esedoglu}.

While our algorithms are inspired by continuous processes, they can
be written directly in a discrete combinatorial setting defined by the graph
Laplacian.  This has the advantage, noted by Grady~\cite{grady06}, of
avoiding errors that could arise from a discretization
process.
We represent the data as nodes in a weighted graph, with each edge assigned a measure of similarity between the vertices it is connecting. The edges between nodes in the graph are not the result of a regular grid embedded in an Euclidean space. Therefore, a nonlocal calculus formulation
\cite{gilboa} is the tool used to generalize the continuous formulation to a (nonlocal) discrete setting given by functions on graphs. Other nonlocal formulations for weighted graphs are included in~\cite{elmoataz:lezoray:bougleux}, while~\cite{grady:polimeni} constitutes a comprehensive reference about techniques to cast continuous PDEs in graph form.The approach of defining functions with domains corresponding to nodes in a graph has successfully been used in areas, such as spectral graph theory \cite{chung, mohar}. 


Graph-based formulations have been used extensively for image processing applications~\cite{shi:malik, boykov:veksler, elmoataz:lezoray:bougleux, grady05, grady06, couprie:grady:najman, couprie:grady:talbot, grady:schiwietz:aharon, levin:acha:lischinski}. 
Interesting connections between these different algorithms, as well as between continuous and discrete optimizations, have been established in the literature. 
Grady has proposed a random walk algorithm~\cite{grady06} that performs
interactive image segmentation using the solution to a combinatorial
Dirichlet problem.  Elmoataz et al.\ have developed
generalizations of the graph Laplacian~\cite{elmoataz:lezoray:bougleux}
for image denoising and manifold smoothing.
Couprie et al. in~\cite{couprie:grady:najman} define a conveniently
parameterized graph-based energy function that is able to unify graph
cuts, random walker, shortest paths and watershed optimizations. There,
the authors test different seeded image segmentation algorithms, and
discuss possibilities to optimize more general models with applications
beyond image segmentation. In~\cite{couprie:grady:talbot}, alternative
graph-based formulations of the continuous max-flow problem are
compared, and it is shown that not all the properties satisfied in the continuous setting carry over to the discrete graph representation. For general data segmentation, Bresson et al. in~\cite{bresson:laurent}, present rigorous convergence results for two algorithms that solve the relaxed Cheeger cut minimization, and show a formula that gives the correspondence between the global minimizers of the relaxed problem and the global minimizers of the combinatorial problem. In our case, 
the convergence property of GL to TV has been known to hold in the
continuum~\cite{kohn}, but has recently been shown in the graph setting
as well~\cite{bertozzi:vangennip}.




\subsection{Binary Segmentation using the Ginzburg-Landau Functional}
\label{review}

The classical Ginzburg-Landau (GL) functional can be written as:
\begin{equation}
             GL(u)=\frac{\epsilon}{2} \int|\nabla u|^{2}dx
+\frac{1}{\epsilon}\int \Phi(u)dx,
\label{eq:GL}
\end{equation}
where $u$ is a scalar field defined over a space of arbitrary
dimensionality and representing the state of the phases in the system,
$\nabla$ denotes the spatial gradient operator, $\Phi(u)$ is a double-well
potential, such as $\Phi(u)=\frac{1}{4}(u^{2}-1)^{2}$, and $\epsilon$ is a
positive constant. The two terms are: a smoothing term
that measures the differences in the components of the field, and a
potential term that measures how far each component is from a specific
value ($\pm1$ in the example above). 
In the next subsection, we derive the proper
formulation in a graph setting.

It is shown in~\cite{kohn} that the $\epsilon \rightarrow 0$ limit of
the GL functional, in the sense of $\Gamma$-convergence, is the Total
Variation (TV) semi-norm: 
\begin{equation}
    GL(u) \rightarrow_\Gamma  || u ||_{TV}.
\end{equation}
Due to this relationship, the two functionals can sometimes be interchanged.
 The advantage of the GL functional is that its $L_2$ gradient flow leads
to a linear differential operator, which allows us to use fast methods for minimization.

 Equation (3) arises in its continuum form in several imaging applications including inpainting~\cite{bertozzi:esedoglu:gillette} and segmentation~\cite{esedoglu}.  In such problems, one typically considers a gradient flow in which the continuum Laplacian is most often discretized in space using the 4-regular graph.  The inpainting application in~\cite{bertozzi:esedoglu:gillette} considers a gradient flow in an $H^{-1}$ inner product resulting in the biharmonic operator which can be discretized by considering two applications of the discrete Laplace operator.  The model in (3) has also been generalized to wavelets~\cite{dobrosotskaya:bertozzi:08,dobrosotskaya:bertozzi:10} by replacing the continuum Laplacian with an operator that has eigenfunctions specified by the wavelet basis.  Here we consider a general graphical framework in which the graph Laplacian replaces the continuum Laplace operator.

We also note that the standard practice in all of the examples above is to introduce an additional term in the energy functional to escape from
trivial steady-state solutions (e.g., all labels taking on
the same value). This leads to the expression
\begin{equation}
 E(u)= GL(u)+ F(u,\hat{u}),
\label{energy}
\end{equation}
where $F$ is the additional term, usually called \emph{fidelity}. This
term allows the specification of any known information, for example, regions of an image that belong to a certain class.

\label{ssl_graphs}
Inspired in part by the PDE-based imaging community, where variational
algorithms combining ideas from spectral methods on graphs with nonlinear edge detection methods are common~\cite{gilboa}, Bertozzi and Flenner extended in~\cite{bertozzi:flenner} the $L_2$ gradient flow of the Ginzburg-Landau (GL) energy functional to the domain of functions on a graph. 

The energy $E(u)$ in (\ref{energy}) can be minimized in the $L_2$ sense using gradient descent. This leads to the following dynamic equation (\emph{modified Allen-Cahn equation}):
\begin{equation}
\frac{\partial u}{\partial t} = - \frac{\delta GL}{\delta u} - \mu
\frac{\delta F}{\delta u} = \epsilon \Delta u - \frac{1}{\epsilon}
\Phi'(u) - \mu \frac{\delta F}{\delta u} \label{eq:mAC}
\end{equation}
where $\Delta$ is the Laplacian operator. A local minimizer is obtained by evolving this expression to steady state. Note
that $E$ is not convex, and may have multiple local minima.

Before continuing further, let us introduce some graph concepts that we will use in subsequent sections.

\subsubsection{Graph Framework for Large Data Sets}
Let $G$ be an undirected graph $G=(V,E)$, where $V$ and $E$ are the sets
of vertices and edges, respectively. The vertices are the building
blocks of the data set, such as points in  $\mathbb{R}^{n}$ or pixels in
an image. The similarity between vertices $i$ and $j$ is measured by a
weight function $w(i,j)$ that satisfies the symmetric property $w(i,j)$=
$w(j,i)$. A large value of $w(i,j)$ indicates that vertices $i$ and $j$
are similar to each other, while a small $w(i,j)$ indicates that they
are dissimilar.  For example, an often used similarity measure is the
Gaussian function
\begin{equation}
\label{eq:gaussian}
        w(i,j)=\exp\left(-\frac{d(i,j)^{2}}{\sigma^2}\right),
\end{equation}
with $d(i,j)$
representing the distance between the points associated with
vertices $i$ and $j$, and $\sigma^2$ a positive parameter. 

Define $\mathbf{W}$ as the matrix $W_{ij}=w(i,j)$, and define the
degree of a vertex $i \in V$ as
\begin{equation}
          d_i= \sum_{j \in V} w(i,j).
\end{equation}
If $\mathbf{D}$ is the diagonal matrix with elements $d_i$, then
the graph Laplacian is defined as the matrix $\mathbf{L=D-W}$.

\subsubsection{Ginzburg-Landau Functional on Graphs}

The continuous GL formulation is generalized to the case of weighted graphs via the graph Laplacian. Nonlocal calculus, such as that outlined in \cite{gilboa},
shows that the Laplace operator is related to the graph
Laplacian matrix defined above, and that the eigenvectors of the
discrete Laplacian converge to the eigenvectors of the Laplacian
\cite{bertozzi:flenner}. However, to guarantee convergence to the
continuum differential operator in the limit of large sample size, the
matrix $\mathbf{L}$ must be correctly scaled
\cite{bertozzi:flenner}. Although several versions exist, we use the symmetric normalized Laplacian
\begin{equation}
\Ls = \mathbf{D}^{-\frac{1}{2}}\mathbf{LD}^{-\frac{1}{2}} =
\mathbf{I} - \mathbf{D}^{-\frac{1}{2}}\mathbf{WD}^{-\frac{1}{2}}. \label{eq:Ls}
\end{equation}
since its symmetric property allows for more
efficient implementations. Note that $\Ls$ satisfies:
 \begin{equation}
      \langle \boldsymbol{u}, \Ls\boldsymbol{u} \rangle = \frac{1}{2} \sum_{i,j} w(i,j) \left ( \frac{u_{i}}{\sqrt{d_i}} - \frac{u_{j}}{\sqrt{d_j}} \right )^{2}
\label{inner}
\end{equation}
for all $\boldsymbol{u} \in \mathbb{R}^{n}$. Here the subscript $i$ refers to the $i^{th}$ coordinate of the vector, and the brackets denote the standard dot product. Note also that $\Ls$ has nonnegative, real-valued eigenvalues.

Likewise, it is important to point out that for tasks such as data classification, the use of a graphs has the advantage of providing a way to deal with nonlinearly separable classes as well as simplifying the processing of high dimensional data. 

The GL functional on graphs is then expressed as
\begin{equation}
GL(\boldsymbol{u}) = \frac{\epsilon}{2} \langle \boldsymbol{u},
\Ls\boldsymbol{u} \rangle +\frac{1}{4 \epsilon}\sum_{i \in V} \left (
u_i^2 - 1 \right )^2, \label{eq:GLF_graphs}
\end{equation}
where $u_i$ is the (real-valued) state of node $i$.  The first term 
replaces the gradient term in (\ref{eq:GL}), and the second term
is the double-well potential,
appropriate for binary classifications.

\subsubsection{Role of Diffuse Interface Parameter $\epsilon$}

In the minimization of the GL functional, two conflicting requirements
are balanced.  The first term tries to maintain a smooth state
throughout the system, 
while the second term tries to force each node to adopt the values
corresponding to the minima of the double-well potential function. 
The two terms are balanced through the diffuse interface parameter $\epsilon$.

Recall that in the continuous case, it is known that the GL functional
(\emph{smoothing} + \emph{potential}) converges to total variation (TV)
in the limit where the diffuse interface parameter $\epsilon \to
0$~\cite{kohn}. An analogous property has recently been shown in the
case of graphs as well, for binary
segmentations~\cite{bertozzi:vangennip}. Since TV is an $L_1$-based
metric, TV-minimization leads to sparse solutions, namely indicator
functions that closely resemble the discrete solution of the original
NP-hard combinatorial segmentation problem~\cite{bresson:laurent:2013,
szlam}. 
Thus, the GL functional actually becomes an $L_1$ metric in the small
$\epsilon$ limit, and leads to sharp transitions between classes.
Intuitively, the convergence of GL to TV holds because in the limit of a vanishing interface, the potential takes precedence and the graph nodes are forced towards the minima of the potential, achieving a configuration of minimal length of transition.
This is contrast to more traditional spectral clustering approaches, which
can be understood as $L_2$-based methods and do not favor sparse solutions.
Furthermore, while the smoothness of the transition in the GL functional
is regulated by $\epsilon$, in practice the value of $\epsilon$ does not
have to be decreased all the way to zero to obtain sharp transitions (an
example of this is shown later in Figure~\ref{fig:3moon_eps}). This capability of modeling the separation of a domain into regions or phases with a controlled smoothness transition between them makes the diffuse interface description attractive for segmentation problems, and distinguishes it from more traditional graph-based spectral partitioning methods. 

\subsubsection{Semi-Supervised Learning (SSL) on Graphs}

In graph-based
learning methods, the graph is constructed such that the edges represent
the similarities in the data set and the nodes have an associated real
state that encodes, with an appropriate thresholding operation, class
membership. 

In addition, in some data sets, the label of a small fraction of data
points is known beforehand. This considerably improves the learning accuracy, explaining in
part the popularity of semi-supervised learning methods. The graph
generalization of the diffuse interface model handles this condition
by using the labels of known points. The GL functional for SSL is:
\begin{eqnarray}
             E(\boldsymbol{u}) & = & \frac{\epsilon}{2} \langle
\boldsymbol{u}, \Ls\boldsymbol{u} \rangle +\frac{1}{4 \epsilon}\sum_{i
\in V} \left ( u_i^2 - 1 \right )^2 \nonumber \\
             & & + \sum_{i \in V} \frac{\mu_i}{2} \left (u_i-\hat{u}_i \right )^{2}. \label{SSL}
\end{eqnarray}
The final term in the sum is the new fidelity term that enforces label values that are known beforehand.  
$\mu_i$ is a parameter that takes the value of a positive constant $\mu$
if $i$ is a fidelity node and zero otherwise, and $\hat{u}_i$ is the
known value of fidelity node $i$. This constitutes a soft assignment of
fidelity points: these are not fixed but allowed to change state.




Note that since GL does not guarantee searching in a space orthogonal to the trivial minimum, alternative constraints could be introduced to obtain partitioning results that do not depend on fidelity information (unsupervised). For example, a mass-balance constraint, $\boldsymbol{u} \perp \boldsymbol{1}$, has been used in~\cite{bertozzi:flenner} to insure solutions orthogonal to the trivial minimum. 

\subsection{MBO Scheme for Binary Classification}

 In \cite{MBO}, Merriman, Bence and Osher propose alternating between the following two steps to approximate motion by mean curvature, or motion in which normal velocity equals mean curvature:

\begin{enumerate}
\item {\em Diffusion.\/} Let $u^{n+\frac{1}{2}}= S(\delta t)u^n$ where
$S(\delta t)$ is the propagator (by time $\delta t$) of the standard heat
equation:
\begin {equation}
       \frac{\partial u}{\partial t} =\Delta u.
\end{equation}
\item {\em Thresholding.\/} Let
\begin{equation*}
u^{n+1}=
  \begin{cases} 1 &\mbox{if } u^{n+\frac{1}{2}} \geq 0, \\
      -1 & \mbox{if } u^{n+\frac{1}{2}} <0.
        \end{cases}
\end{equation*} 
\end{enumerate}
This MBO scheme has been rigorously proven to approximate motion by mean curvature by Barles \cite{barles} and Evans \cite{evans} .

The algorithm is related to solving the basic (unmodified) Allen-Cahn
equation, namely equation (\ref{eq:mAC}) without the fidelity term.
If we consider a time-splitting scheme (details in \cite{esedoglu}) to evolve the equation, in the $\epsilon \rightarrow 0$ limit, the second step
is simply thresholding~\cite{MBO}. Thus, as $\epsilon \rightarrow 0$,
the time splitting scheme above consists of alternating between
diffusion and thresholding steps (MBO scheme mentioned above). 
In fact, it has been shown \cite{rubinstein} that in the limit $\epsilon
\rightarrow 0$, the rescaled solutions $u_{\epsilon}(z,t/\epsilon)$ of
the Allen-Cahn equation yield motion by mean curvature of the interface
between the two phases of the solutions, which the MBO scheme approximates.  

The motion by mean curvature of the scheme can be generalized to the case of functions on a graph in much the same way as the procedure followed for the modified Allen-Cahn equation (\ref{eq:mAC}) in~\cite{bertozzi:flenner}. Merkurjev et al. have pursued this idea in~\cite{merkurjev}, where a modified MBO scheme on graphs has been applied to the case of binary segmentation. The motivation comes from \cite{esedoglu} by Esedoglu and Tsai, who propose threshold dynamics for the two-phase piecewise constant Mumford-Shah (MS) functional. The authors derive the scheme by applying a two-step time splitting scheme to the gradient descent equation resulting from the minimization of the MS functional, so that the second step is the same as the one in the original MBO scheme. Merkurjev et al. in \cite{merkurjev} also apply a similar time splitting scheme, but now to  (\ref{eq:mAC}). The $\Delta u$ term is then replaced with a more general graph term
$-\Ls\boldsymbol{u}$. The discretized version of the algorithm is:
\begin{enumerate}
\item Heat equation with forcing term:
\begin{equation}
             \frac{\boldsymbol{u}^{n+\frac{1}{2}}- \boldsymbol{u}^{n}}{dt}= - \Ls
\boldsymbol{u}^{n}-
\boldsymbol{\mu}(\boldsymbol{u}^{n}-\boldsymbol{\hat{u}}).
\label{eq:MBO_step1_binary}
\end{equation}
\item Thresholding:
	\begin{equation*}
                u^{n+1}_{i} =
                \begin{cases}
                     1, & \text{if $u^{n+\frac{1}{2}}_{i}> 0$}, \\
                     -1, & \text{if $u^{n+\frac{1}{2}}_{i} < 0$}.
                \end{cases}
            \end{equation*}
\end{enumerate}
Here, after the second step, $u^n_i$ can take only two values of $1$ or $-1$; thus, this method is appropriate for binary segmentation.
The fidelity term scaling can be different from the one in (\ref{eq:mAC}). 

	The following section describes the modifications introduced to generalize this functional to multiclass segmentation.

\section{Multiclass Data Segmentation}
\label{first_algorithm}

The main point of this paper is to show how to extend prior work to the multiclass case. This allows us to tackle a broad class of machine learning problems.

We use the following notation in the multiclass case. Given $N_D$ data points, we generalize the label vector $\boldsymbol{u}$ to a label matrix
$\mathbf{U} = (\mathbf{u}_1, \dots, \mathbf{u}_{N_D})^T$.  Rather than node $i$ adopting a single state
$u_i \in \mathbb{R}$, it now adopts a composition of states expressed by a vector
$\mathbf{u}_i\in \mathbb{R}^K$ where the $k$th component of $\mathbf{u}_i$ is the strength with which
it takes on class $k$.  The matrix $\mathbf{U}$ has dimensions
$N_D \times K$, where $K$ is the total number of possible classes.

For each node $i$, we require the vector $\mathbf{u}_i$ to be an element of the Gibbs simplex $\Sigma^K$, defined as
\begin{equation}
	 \Sigma^K := \left \{ (x_1, \dots, x_K) \in [0,1]^K \,\middle | \,
\sum_{k=1}^K x_{k} = 1 \right \}.
\end{equation}
Vertex $k$ of the simplex is given by the unit vector
$\boldsymbol{e}_k$, whose $k$th component equals 1 and all
other components vanish.  These vertices correspond to pure phases, where 
the node belongs exclusively to class $k$.  The simplex
formulation has a probabilistic interpretation, with 
$\mathbf{u}_i$ representing the probability distribution over the $K$ classes.
In other segmentation algorithms, such as spectral clustering, these
real-valued variables can have different interpretations that are exploited
for specific applications, as discussed
in~\cite{grady:schiwietz:aharon,levin:acha:lischinski}.


\subsection{Multiclass Ginzburg-Landau Approach}

The multiclass GL energy functional for the phase field approach on graphs is written as:
\begin{eqnarray}
	E(\mathbf{U}) & = & \frac{\epsilon}{2} \langle
\mathbf{U}, \Ls\mathbf{U} \rangle +
\frac{1}{2 \epsilon}\sum_{i \in V}\left( \prod_{k=1}^K \frac{1}{4} \left
\Vert \Uivec - \boldsymbol{e}_k \right \Vert_{L_1}^2\right) \nonumber \\
	& & + \sum_{i \in V} \frac{\mu_i}{2} \; \left \Vert \Uivec -
\hat{\mathbf{u}}_{i} \right \Vert^2, \label{eq:MGL_SSL}
\end{eqnarray}
where
\begin{eqnarray*}
	\langle \mathbf{U}, \Ls\mathbf{U} \rangle &=&
        \mathrm{trace}( \mathbf{U}^T \Ls\mathbf{U}),
\end{eqnarray*}
and $\mathbf{\hat{u}}_i$ is a vector indicating prior class knowledge of sample $i$.
We set $\boldsymbol{\hat{u}}_i = \boldsymbol{e}_k$ if node $i$ is known to be in
class $k$.

The first (smoothing) term in the GL functional (\ref{eq:MGL_SSL}) measures
variations in the vector field.  The simplex representation has the
advantage that, like in Potts-based
models but unlike in some other multiclass methods, the penalty assigned
to differently labeled neighbors is independent of the integer ordering
of the labels.
The second (potential) term drives
the system closer to the vertices of the simplex.  For this term, we
adopt an $L_1$ norm to prevent
the emergence of an undesirable minimum at
the center of the simplex, as would occur with an $L_2$ norm for large
$K$.  The third (fidelity) term
enables the encoding of \textit{a priori} information.

Note that one can obtain meaningful results without fidelity information
(unsupervised), but the methods for doing so are
not as straightforward.  One example is a new TV-based modularity
optimization method~\cite{hu:laurent:mason} that makes no assumption as
to the number of classes and can be recast as GL minimization. 
Also, while $\Gamma$-convergence to TV in the graph setting has been
proven for the binary segmentation problem~\cite{bertozzi:vangennip}, no similar convergence property has yet
been proven for the multiclass case.  We leave
this as an open conjecture.

Following~\cite{bertozzi:flenner}, we use a convex splitting scheme to minimize the GL functional in the phase field approach. The energy functional (\ref{eq:MGL_SSL}) is decomposed into convex and concave parts:
\begin{eqnarray*}
E(\mathbf{U}) & = & E_{\mathrm{convex}}(\mathbf{U})
+ E_{\mathrm{concave}}(\mathbf{U}) \\
E_{\mathrm{convex}}(\mathbf{U}) & = & \frac{\epsilon}{2}
\langle \mathbf{U}, \Ls\mathbf{U}
\rangle + \frac{C}{2} \langle \mathbf{U}, \mathbf{U} \rangle \\
E_{\mathrm{concave}}(\mathbf{U}) & = & \frac{1}{2
\epsilon}\sum_{i \in V} \prod_{k=1}^K \frac{1}{4} \left \Vert
\Uivec - \boldsymbol{e}_k \right \Vert_{L_1}^2 \\
& & + \sum_{i \in V} \frac{\mu_i}{2} \; \left \Vert \Uivec-
\mathbf{\hat{u}}_i \right \Vert^2_{L_2} - \frac{C}{2} \langle \mathbf{U},
\mathbf{U} \rangle
\end{eqnarray*}
\noindent with $C \in \mathbb{R}$ denoting a constant that is chosen to
guarantee the convexity/concavity of the energy terms. Evaluating the second derivative of the partitions, and simplifying terms, yields:
\begin{equation}
	C \geq \mu + \frac{1}{\epsilon} \label{eq:limitC} .
\end{equation}

The convex splitting scheme results in an unconditionally stable time-discretization scheme using a gradient descent implicit in the convex partition and explicit in the concave partition, as given by the form~\cite{eyre,yuille:rangarajan,esedoglu} 
\begin{equation}
U_{ik}^{n+1} + dt \, \frac{\delta E_{\mathrm{convex}}}{\delta
U_{ik}} (U_{ik}^{n+1}) =
U_{ik}^n
- dt \, \frac{\delta E_{\mathrm{concave}}}{\delta U_{ik}}
(U_{ik}^{n}).
\end{equation}
We write this equation in
matrix form as
\begin{eqnarray}
&& \hspace{-0.25in}\mathbf{U}^{n+1} + dt\left(\epsilon\Ls\mathbf{U}^{n+1} + C
\mathbf{U}^{n+1}\right) \nonumber  \\
&& = \mathbf{U}^n - dt
\left(\frac{1}{2\epsilon}\mathbf{T}^n + \boldsymbol{\mu}
(\mathbf{U}^n-\mathbf{\hat{U}}) - C \mathbf{U}^n\right),
\label{eq:semi_GD}
\end{eqnarray}
where
\begin{equation}
T_{ik} = \sum_{l=1}^K \frac{1}{2} \left ( 1 - 2
\delta_{k l} \right ) \left \Vert \Uivec -
\boldsymbol{e}_l \right \Vert_{L_1} 
\prod_{\substack{m=1 \\ m\neq l}}^K
\frac{1}{4} \left \Vert \Uivec - \boldsymbol{e}_m \right
\Vert_{L_1}^2,
\end{equation}
$\boldsymbol{\mu}$ is a diagonal matrix with elements $\mu_i$, and $\mathbf{\hat{U}} = (\mathbf{\hat{u}}_1, \dots, \mathbf{\hat{u}}_{N_D})^T$.

Solving (\ref{eq:semi_GD}) for $\mathbf{U}^{n+1}$ gives the iteration
equation
\begin{equation}
\mathbf{U}^{n+1} = \mathbf{B}^{-1}
\left[ (1 + C \; dt )
\,\mathbf{U}^n
- \frac{dt}{2 \epsilon} \, \mathbf{T}^n
 - dt \,
\boldsymbol{\mu}(\mathbf{U}^n-\mathbf{\hat{U}})\right]
\label{eq:updateGL}
\end{equation}
where
\begin{equation}
\mathbf{B} =  (1+C\,dt)\mathbf{I} +
\epsilon\,dt\,\Ls.
\end{equation}
This implicit scheme allows the evolution of $\mathbf{U}$ to be
numerically stable regardless of the time step $dt$, in spite of the
numerical ``stiffness'' of the underlying differential equations which
could otherwise force $dt$ to be impractically small.

In general, after the update, the phase field is no longer on the
$\Sigma^K$ simplex. Consequently, we use the procedure in~\cite{chen}
to project back to the simplex.

Computationally, the scheme's numerical efficiency is increased by
using a low-dimensional subspace spanned by only a small
number of eigenfunctions.
Let $\mathbf{X}$ be the matrix of eigenvectors of $\Ls$ and 
$\boldsymbol{\Lambda}$ be the diagonal matrix of corresponding
eigenvalues.
We now write $\Ls$ as its eigendecomposition
$\Ls=\mathbf{X}\boldsymbol{\Lambda}\mathbf{X}^T$, and set
\begin{equation}
\mathbf{B}=\mathbf{X} \left[(1+C\,dt)\mathbf{I} + \epsilon\,dt\,\boldsymbol{\Lambda}\right] \mathbf{X}^T,
\end{equation}
but we approximate $\mathbf{X}$ by a truncated matrix retaining only
$N_e$ eigenvectors ($N_e \ll N_D$), to form a matrix of dimension $N_D
\times N_e$.  The term in brackets is simply a diagonal $N_e \times N_e$
matrix. This allows $\mathbf{B}$ to be calculated rapidly, but more
importantly it allows the update step (\ref{eq:updateGL}) to be decomposed
into two significantly faster matrix multiplications (as discussed
below), while sacrificing little accuracy in practice.

\begin{figure*}[ht]
\caption{Multiclass GL Algorithm} \label{algo:iter_MGL}
\rule{18cm}{0.2mm}
\begin{algorithmic} 
\REQUIRE $\epsilon, dt, N_D, N_e, K, \boldsymbol{\mu}, \mathbf{\hat{U}},
\boldsymbol{\Lambda},\mathbf{X}$
\ENSURE $\mathrm{out} = \mathbf{U^{end}}$
\STATE $C \leftarrow \mu + \frac{1}{\epsilon}$
\STATE $\mathbf{Y} \leftarrow \left[(1 + C\,dt)\mathbf{I} +
\epsilon\,dt\boldsymbol{\Lambda}\right]^{-1}\mathbf{X}^T$
\FOR{$i = 1 \to N_D$}
\STATE $U_{i k}^{~0} \leftarrow rand((0,1)), \
U_{ik}^0 \leftarrow projectToSimplex(\Uivec^0).
\quad \mathrm{If~} \mu_i > 0, ~ U_{i k}^{~0} \leftarrow
\hat{U}_{i k}^{~0}$
\ENDFOR
\STATE $n \leftarrow 1$
\WHILE{$\mathrm{Stop~criterion~not~satisfied}$}
\FOR{$i =1 \to N_D$, $k =1 \to K$}
\STATE $T_{i k}^{~n} \leftarrow \sum_{l=1}^K
\frac{1}{2} \left ( 1 - 2 \delta_{k l} \right ) \left \Vert
\Uivec^n - \boldsymbol{e}_l \right \Vert_{L_1}
\prod_{m=1,
m \neq l}^K \frac{1}{4} \left \Vert \Uivec^n -
\boldsymbol{e}_m \right \Vert_{L_1}^2 $
\ENDFOR
\STATE $\mathbf{Z} \leftarrow \mathbf{Y}
\left [ (1 + C \; dt) \,\mathbf{U}^n
- \frac{dt}{2 \epsilon} \, \mathbf{T}^n - dt \,
\boldsymbol{\mu}(\mathbf{U}^n-\mathbf{\hat{U}}) \right ] $
\STATE $\mathbf{U}^{n+1} \leftarrow \mathbf{XZ}$
\FOR{$i =1 \to N_D$}
\STATE $\Uivec^{n+1} \leftarrow
projectToSimplex(\Uivec^{n+1})$
\ENDFOR
\STATE $ n \leftarrow n + 1$
\ENDWHILE
\end{algorithmic}
\rule{18cm}{0.2mm}
\end{figure*}

For initialization, the phase compositions of the fidelity points are
set to the vertices of the simplex corresponding to the known labels,
while the phase compositions of the rest of the points are set randomly.

The energy minimization proceeds until a steady state condition is reached. The final classes are obtained by assigning class $k$ to node $i$ if $\Uivec$ is closest to vertex $\boldsymbol{e}_k$ on the Gibbs
simplex. Consequently, the calculation is stopped when
\begin{equation}
\frac{ \max_i \Vert \Uivec^{n+1} - \Uivec^{n} \Vert^{2}}{ \max_i
\Vert \Uivec^{n+1} \Vert^{2}} < \eta, \label{eq:stop}
\end{equation}
where $\eta$ represents a given small positive constant.

The algorithm is outlined in Figure~\ref{algo:iter_MGL}. While
other operator splitting methods have been studied for minimization
problems (e.g.~\cite{Lellmann2008}), ours has the following
advantages: (i) it is direct (i.e. it does not require the solution of
further minimization problems), (ii) the resolution can be adjusted by
increasing the number of eigenvectors $N_e$ used in the representation
of the phase field, and (iii) it has low complexity.  To see this final
point, observe that each iteration of
the multiclass GL algorithm has only
$O(N_D K N_e)$ operations for the main loop, since matrix $\mathbf{Z}$ in
Figure~\ref{algo:iter_MGL} only has dimensions $N_e \times K$, and then
$O(N_D K \log K)$ operations for the projection to the simplex.
Usually, $N_e \ll N_D$ and $K \ll N_D$, so the dominant factor is simply
the size of the data set $N_D$.
In addition, it is generally the case that the number of
iterations required for convergence is moderate (around 50 iterations).
Thus, practically speaking, the complexity of the algorithm is
linear.

\subsection{Multiclass MBO Reduction}
\label{second_algorithm}

\begin{figure*}[ht]
\caption{Multiclass MBO Algorithm} \label{algo:iter_MBO}
\rule{18cm}{0.2mm}
\begin{algorithmic} 
\REQUIRE $dt, N_D, N_e, N_S, K, \boldsymbol{\mu}, \mathbf{\hat{U}},
\boldsymbol{\Lambda},\mathbf{X}$
\ENSURE $\mathrm{out} = \mathbf{U^{end}}$
\STATE $\mathbf{Y} \leftarrow \left(\mathbf{I} +
\frac{dt}{N_S}\boldsymbol{\Lambda}\right)^{-1}\mathbf{X}^T$
\FOR{$i = 1 \to N_D$}
\STATE $U_{i k}^{~0} \leftarrow rand((0,1)), \
\Uivec^0 \leftarrow projectToSimplex(\Uivec^0).
\quad \mathrm{If~} \mu_i > 0, ~ U_{i k}^{~0} \leftarrow
\hat{U}_{i k}^{~0}$
\ENDFOR
\STATE $n \leftarrow 1$
\WHILE{$\mathrm{Stop~criterion~not~satisfied}$}
\FOR{$s=1 \to N_S$}
\STATE $\mathbf{Z} \leftarrow \mathbf{Y}
\left [ \mathbf{U}^n - \frac{dt}{N_S} \,
\boldsymbol{\mu}(\mathbf{U}^n-\mathbf{\hat{U}}) \right ] $
\STATE $\mathbf{U}^{n+1} \leftarrow \mathbf{X}\mathbf{Z}$
\ENDFOR
\FOR{$i =1 \to N_D$}
\STATE $\Uivec^{n+1} \leftarrow
projectToSimplex(\Uivec^{n+1})$
\STATE $\Uivec^{n+1} \leftarrow
\boldsymbol{e}_k$, where $k$ is closest simplex vertex to
$\Uivec^{n+1}$
\ENDFOR
\STATE $ n \leftarrow n + 1$
\ENDWHILE
\end{algorithmic}
\rule{18cm}{0.2mm}
\end{figure*}

Using the standard Gibbs-simplex $\Sigma^K$, the multiclass extension of the algorithm in
\cite{merkurjev} is straightforward. The notation is the same as in the beginning of the section.
While the first step of the algorithm remains the same (except, of course, it is now in matrix form), the second step of the algorithm is modified so that the
thresholding is converted to the displacement of the vector field
variable towards the closest vertex in the Gibbs simplex. In other
words, the row vector $\Uivec^{n+\frac{1}{2}}$
of step $1$ is projected back to the simplex (using the approach
outlined in \cite{chen} as before) and then a pure phase given
by the vertex in the $\Sigma^K$ simplex closest to
$\Uivec^{n+\frac{1}{2}}$ 
is assigned to be the new phase composition of node $i$.

In summary, the new algorithm consists of alternating between the following two steps to obtain approximate solutions $\mathbf{U}^{n}$ at discrete times: 
\begin{enumerate}
\item Heat equation with forcing term:
\begin{equation}
             \frac{\mathbf{U}^{n+\frac{1}{2}}- \mathbf{U}^{n}}{dt}= -
\Ls \mathbf{U}^{n}-
\boldsymbol{\mu}(\mathbf{U}^{n}-\mathbf{\hat{U}}).
\label{eq:MBO_step1_multiclass}
\end{equation}
\item Thresholding: 
\begin{equation}
               \Uivec^{n+1}= \boldsymbol{e}_k,
\end{equation}
where vertex $\boldsymbol{e}_k$ is the vertex in the simplex
closest to $projectToSimplex(\Uivec^{n+\frac{1}{2}})$. 
\end{enumerate}
As with the multiclass GL algorithm, when a label is known, it is
represented by the corresponding vertex in the $\Sigma^K$ simplex. The
final classification is achieved by assigning node $i$ to class
$k$ if if the $k$th component of $\mathbf{u}_i$ is one. Again, as in the binary case, the diffusion step can be repeated a number of times before thresholding and when that happens, $dt$ is divided by the number of diffusion iterations $N_S$.

As in the previous section, we use an implicit numerical scheme.  For the MBO algorithm, the procedure involves modifying
(\ref{eq:MBO_step1_multiclass}) to apply $\Ls$ to
$\mathbf{U}^{n+\frac{1}{2}}$ instead of to $\mathbf{U}^n$.
This gives the diffusion step
\begin{equation}
\mathbf{U}^{n+\frac{1}{2}} = \mathbf{B}^{-1} \left[\mathbf{U}^n
 - dt \,
\boldsymbol{\mu}(\mathbf{U}^n-\mathbf{\hat{U}})\right]
\end{equation}
where
\begin{equation}
\mathbf{B}= \mathbf{I} + dt\,\Ls.
\end{equation}
As before, we use the eigendecomposition
$\Ls=\mathbf{X}\boldsymbol{\Lambda}\mathbf{X}^T$ to write
\begin{equation}
\mathbf{B}=\mathbf{X}\left(\mathbf{I} + dt\,\boldsymbol{\Lambda}\right)
\mathbf{X}^T,
\end{equation}
which we approximate using the first $N_e$ eigenfunctions.
The initialization procedure and the stopping criterion are the same as
in the previous section.


The multiclass MBO algorithm is summarized in Figure~\ref{algo:iter_MBO}.
Its complexity is $O(N_D K N_e N_S)$ operations for the main loop,
$O(N_D K \log K)$ operations for the projection to the simplex and
$O(N_D K)$ operations for thresholding.  As in the multiclass GL
algorithm, $N_e \ll N_D$ and $K \ll N_D$.  Furthermore, $N_S$ needs to be set to three, and due to the thresholding step, we find that extremely few
iterations (e.g., 6) are needed to reach steady state.  Thus, in
practice, the complexity of this algorithm is linear as well, and
typical runtimes are very rapid as shown in Table \ref{timing}.

Note that graph analogues of continuum operators, such as gradient and Laplacian, can be constructed using tools of nonlocal discrete calculus. Hence, it is possible to express notions of graph curvature for arbitrary graphs, even with no geometric embedding, but this is not straightforward. For a more detailed discussion about the MBO scheme and motion by mean curvature on graphs, we refer the reader to~\cite{vangennip:guillen:osting}.

\section{Experimental Results}
\label{results}

We have tested our algorithms on synthetic data, image labeling, and the
MNIST, COIL and WebKB benchmark data sets. In most of these cases,
we compute the symmetric normalized graph Laplacian matrix $\Ls$, of
expression (\ref{eq:Ls}), using $N$-neighborhood graphs: in other words,
vertices $i$ and $j$ are connected only if $i$ is among the $N$ nearest
neighbors of $j$ or if $j$ is among the $N$ nearest neighbors of $i$.
Otherwise, we set $w(i,j)=0$. This results in a sparse matrix, making
calculations and algorithms more tractable. In addition, for the
similarity function we use the local scaling weight function of
Zelnik-Manor and Perona~\cite{zelnik-manor}, defined as
\begin{equation}
        w(i,j)=\exp\left(-\frac{d(i,j)^{2}}{\sqrt{\tau(i)\tau(j)}}\right)
\end{equation}
where $d(i,j)$ is some distance measure between vertices $i$ and $j$,
such as the $L_{2}$ distance, and $\sqrt{\tau(i)}= d(i,k)$ defines a
local value for each vertex $i$, parametrized by $M$, with $k$ being the
index of the $M$th closest vertex to $i$.

With the exception of the image labeling example, all the results and comparisons
with other
published methods are summarized in Tables \ref{benchmark} and
\ref{webkbfidelity}. Due to the arbitrary selection of the fidelity
points, our reported values correspond to averages obtained over 10 runs
with different random selections. The timing results and number of
iterations of the two methods are shown in Tables \ref{timing} and
\ref{iterations}, respectively. The methods are labeled as ``multiclass
GL'' and ``multiclass MBO''. These comparisons show that our methods
exhibit a performance that is competitive with or better than the current state-of-the-art segmentation algorithms.

Parameters are chosen to produce comparable performance between the methods. For the multiclass GL method, the convexity constant used is: $C = \mu + \frac{1}{\epsilon}$. As described before in expression (\ref{eq:limitC}), this is the lower limit that guarantees the convexity and concavity of the terms in the energy partition of the convex splitting strategy employed. For the multiclass MBO method, as discussed in the previous section, the diffusion step can be repeated a number of times before thresholding. In all of our results, we run the diffusion step three times before any thresholding is done ($N_S = 3$).

To compute the eigenvectors and eigenvalues of the symmetric graph
Laplacian, we use fast numerical solvers. As we only need to calculate a
portion of the eigenvectors to get good results, we compute the
eigendecompositions using the
Rayleigh-Chebyshev procedure of \cite{anderson}
in all cases except the image labeling example. This numerical solver is
especially efficient for
producing a few of the smallest eigenvectors of a sparse symmetric
matrix. For example, for the MNIST data set of 70,000 images, it was
only necessary to calculate $300$ eigenvectors, which is less than
$0.5\%$ of the data set size. This is one of the factors that makes our
methods very efficient. For the image labeling experiments, we use the Nystr\"{o}m extension method described in~\cite{bertozzi:flenner,fowlkes:belongie:malik,fowlkes:belongie:chung}. The advantage of the latter method is that it can be efficiently used for very large datasets, because it appoximates the eigenvalues and eigenvectors of a large matrix by calculations done on much smaller matrices formed by randomly chosen parts of the original matrix.

\begin{table*}[!t]
\caption{Results for benchmark data sets: Moons, MNIST, COIL and WebKB}
\label{benchmark}
\begin{center}
\begin{subtable}{1.8 in} 
\caption*{\fontsize{11}{9}\selectfont {Two/Three moons}}
\begin{tabular}{|c| c|}
\hline
Method & Accuracy \\
\hline
spectral clustering~\cite{gc:flenner:percus} & 80\% \\
\hline
p-Laplacian~\cite{buhler} & 94\% \\
\hline 
Cheeger cuts~\cite{szlam} & 95.4\% \\
\hline
tree GL~\cite{gc:flenner:percus} & 97.4\% \\
\hline
binary GL~\cite{bertozzi:flenner} & 97.7\% \\
\hline
{\em multiclass GL\/} & 98.1\% \\
\hline
{\em multiclass MBO\/} & 99.12\% \\
\hline
\end{tabular}
\end{subtable}
\hspace{0.35in}
\begin{subtable}{2.6 in}
\caption*{\fontsize{11}{9}\selectfont {MNIST}}
\begin{tabular}{|c| c|}
\hline
Method & Accuracy \\
\hline
p-Laplacian~\cite{buhler} & 87.1\% \\
\hline
multicut normalized 1-cut~\cite{hein} & 87.64\%\\
\hline
linear classifiers~\cite{lecun,lecun:cortes} & 88\% \\
\hline
Cheeger cuts~\cite{szlam} & 88.2\% \\
\hline
boosted stumps~\cite{kegl,lecun:cortes} & 92.3-98.74\% \\
\hline
transductive classification~\cite{szlam:maggioni:coifman} & 92.6\% \\
\hline
tree GL~\cite{gc:flenner:percus} & 93.0\% \\
\hline
$k$-nearest neighbors~\cite{lecun,lecun:cortes} & 95.0-97.17\% \\
\hline
neural/convolutional nets~\cite{lecun,ciresan,lecun:cortes} & 95.3-99.65\% \\
\hline
nonlinear classifiers~\cite{lecun,lecun:cortes} & 96.4-96.7\% \\
\hline
{\em multiclass GL\/} & 96.8\%  \\
\hline
{\em multiclass MBO\/} & 96.91\% \\
\hline
SVM~\cite{lecun,decoste} & 98.6-99.32\%  \\
\hline
\end{tabular}
\end{subtable}
\end{center}

\vspace{0.1in}

\begin{center}
\begin{subtable}{1.9 in}
\caption*{\fontsize{11}{9}\selectfont {COIL}}
\begin{tabular}{|c| c|}
\hline
Method & Accuracy \\
\hline
$k$-nearest neighbors~\cite{subramanya} & 83.5\% \\
\hline
LapRLS~\cite{belkin,subramanya} & 87.8\% \\
\hline
sGT~\cite{joachims,subramanya} & 89.9\% \\
\hline
SQ-Loss-I~\cite{subramanya} & 90.9\% \\
\hline
MP~\cite{subramanya} & 91.1\% \\
\hline
{\em multiclass GL\/} & 91.2\% \\
\hline
{\em multiclass MBO\/} & 91.46\% \\
\hline
\end{tabular}
\end{subtable}
\hspace{0.35in}
\begin{subtable}{2.3 in}
\caption*{\fontsize{11}{9}\selectfont {WebKB}}
\begin{tabular}{|c| c|}
\hline
Method & Accuracy \\
\hline
vector method~\cite{cardoso} & 64.47\% \\
\hline
$k$-nearest neighbors ($k=10$)~\cite{cardoso} & 72.56\% \\
\hline
centroid (normalized sum)~\cite{cardoso} & 82.66\% \\
\hline
naive Bayes~\cite{cardoso} & 83.52\% \\
\hline
SVM (linear kernel)~\cite{cardoso} & 85.82\% \\
\hline
{\em multiclass GL\/} & 87.2\% \\
\hline
{\em multiclass MBO\/} & 88.48\% \\
\hline
\end{tabular}
\end{subtable}
\end{center}
\end{table*}

\begin{table*}[!t]
\centering
\caption{WebKB results with varying fidelity percentage}
\label{webkbfidelity}

\renewcommand{\arraystretch}{1.3}

\begin{tabular}{|c|| c|c| c| c| c|}
\hline
Method & 10\% & 15\% & 20\% & 25\% & 30\% \\
\hline
WebKB results for Multiclass GL (\% correct) & 81.3\% & 84.3\% & 85.8\% & 86.7\% & 87.2\% \\
\hline
WebKB results for Multiclass MBO (\% correct) & 83.71\% & 85.75\% & 86.81\% & 87.74\% & 88.48\%\\
\hline
\end{tabular}
\end{table*}

\begin{table*}[!t]
\centering
\caption{Comparison of timings (in seconds)}
\label{timing}
\renewcommand{\arraystretch}{1.3}
\begin{tabular}{|c|| c| c| c| c| c|}
\hline
Data set & three moons & color images & MNIST & COIL & WebKB \\
\hline
Size & 1.5 K & 144 K & 70 K & 1.5 K & 4.2 K \\
\hline
Graph Calculation & 0.771 & 0.52 &6183.1 & 0.95 & 399.35 \\
\hline
Eigenvector Calculation & 0.331 & 27.7 & 1683.5 & 0.19 & 64.78 \\
\hline
Multiclass GL & 0.016 & 837.1 & 153.1 & 0.035 & 0.49 \\
\hline
Multiclass MBO & 0.013 & 40.0 & 15.4 & 0.03 & 0.05 \\
\hline
\end{tabular}
\end{table*}

\begin{table*}[!t]
\centering
\caption{Comparison of number of iterations}
\label{iterations}
\renewcommand{\arraystretch}{1.3}
\begin{tabular}{|c|| c| c| c| c| c|}
\hline
Data set & three moons & color images & MNIST & COIL & WebKB \\
\hline
Multiclass GL & 15 & 770 & 90 & 12 & 20 \\
\hline
Multiclass MBO  & 3 & 44 & 7 & 6 & 7\\
\hline
\end{tabular}
\end{table*}

\subsection{Synthetic Data}
The synthetic data set we tested our method against is the three moons
data set. It is constructed by generating three half circles in
$\mathbb{R}^{2}$. The two half top circles are unit circles with centers
at $(0,0)$ and $(3,0)$. The bottom half circle has radius $1.5$ and the
center at $(1.5,0.4)$. Five hundred points from each of those three half
circles are sampled and embedded in $\mathbb{R}^{100}$ by adding
Gaussian noise with standard deviation of $0.14$ to each of the $100$
components of each embedded point.  The dimensionality of the data set,
together with the noise, makes segmentation a significant challenge.

The weight matrix of the graph edges was calculated using $N = 10$ nearest neighbors and local scaling based on the $17^{th}$ closest point ($M=17$). The fidelity term was constructed by labeling $25$ points per class, $75$ points in total, corresponding to only $5$\% of the points in the data set.

 The multiclass GL method used the following parameters: 
 $15$ eigenvectors, $\epsilon=1$, $dt=0.1$, $\mu=30$, $\eta=10^{-7}$. The method was able to produce an average of 98.1\% of correct classification, with a corresponding computation time of $0.016$ s per run on a $2.4$ GHz Intel Core i2 Quad without any parallel processing.

\begin{figure}[h]
\centering
     \scalebox{0.5} { \includegraphics[clip,width=3.5in]{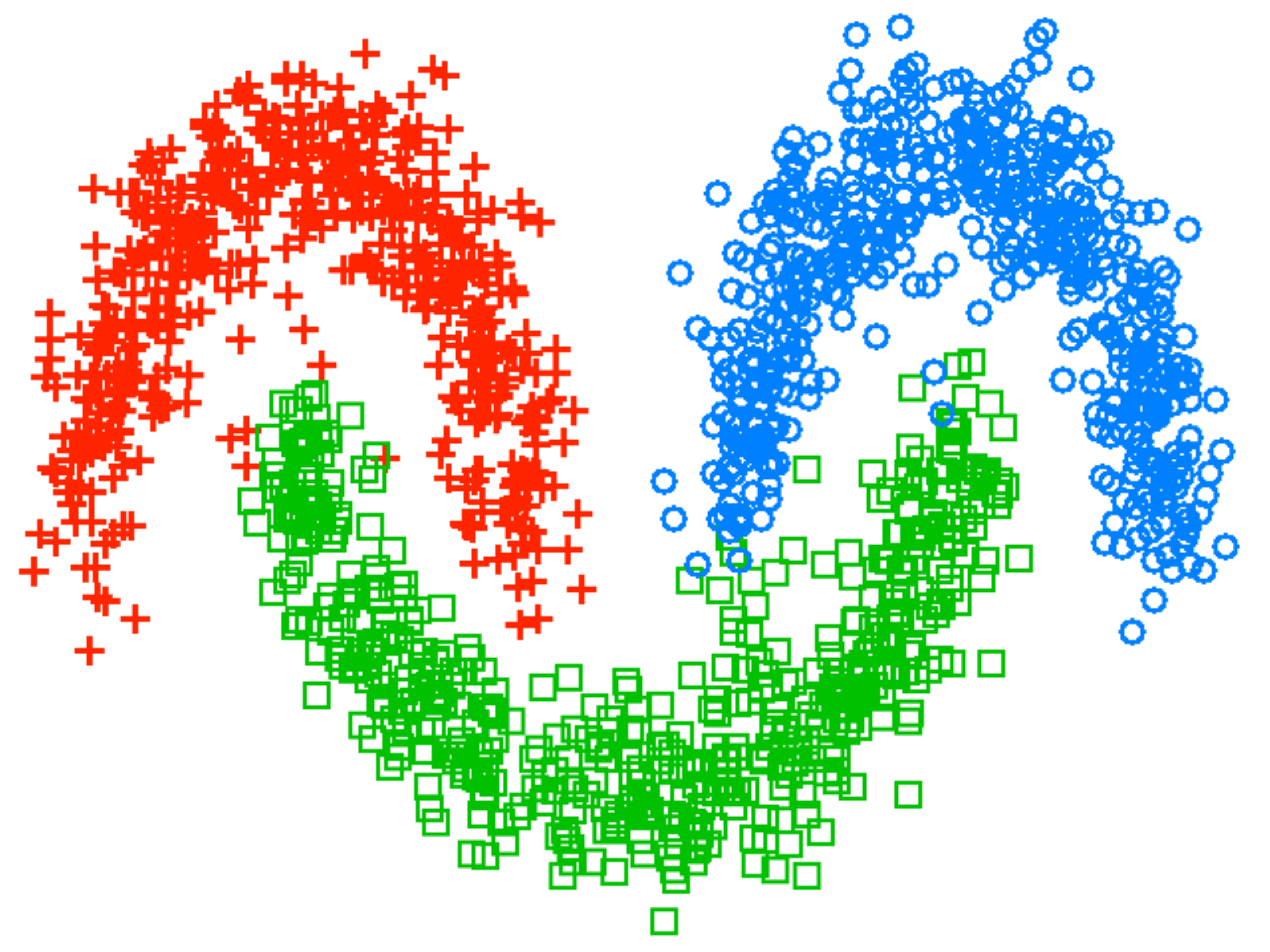}}
 \caption{Segmentation of three moons using multiclass MBO (98.4667\% correct).}
\label{fig:3moons_result}
\end{figure}

Analogously, the multiclass MBO method used the following parameters:
$20$ eigenvectors, $dt=0.1$, $\mu=30$, $\eta=10^{-7}$. It was able to segment an average of $99.12$\% of the points correctly over $10$ runs with only $3$ iterations and about $0.01$ s of computation time. One of the results obtained is shown in Figure \ref{fig:3moons_result}.  

Table \ref{benchmark} gives published results from other related
methods, for comparison.
Note that the results for p-Laplacians~\cite{buhler}, Cheeger
cuts~\cite{szlam} and binary GL are for the simpler binary problem of
two moons (also embedded in $\mathbb{R}^{100}$).  While, strictly speaking,
these are unsupervised methods, they all incorporate prior knowledge such as a
mass balance constraint.  We therefore consider them comparable to our
SSL approach.
The ``tree GL'' method~\cite{gc:flenner:percus} uses a scalar
multiclass GL approach with a tree metric.
It can be seen that our methods achieve the highest accuracy on this
test problem.

\begin{figure}[h!]
         \begin{subfigure}[b]{0.24\textwidth}
                \centering
                     \scalebox{0.4}{\includegraphics[clip,width=3.5in]{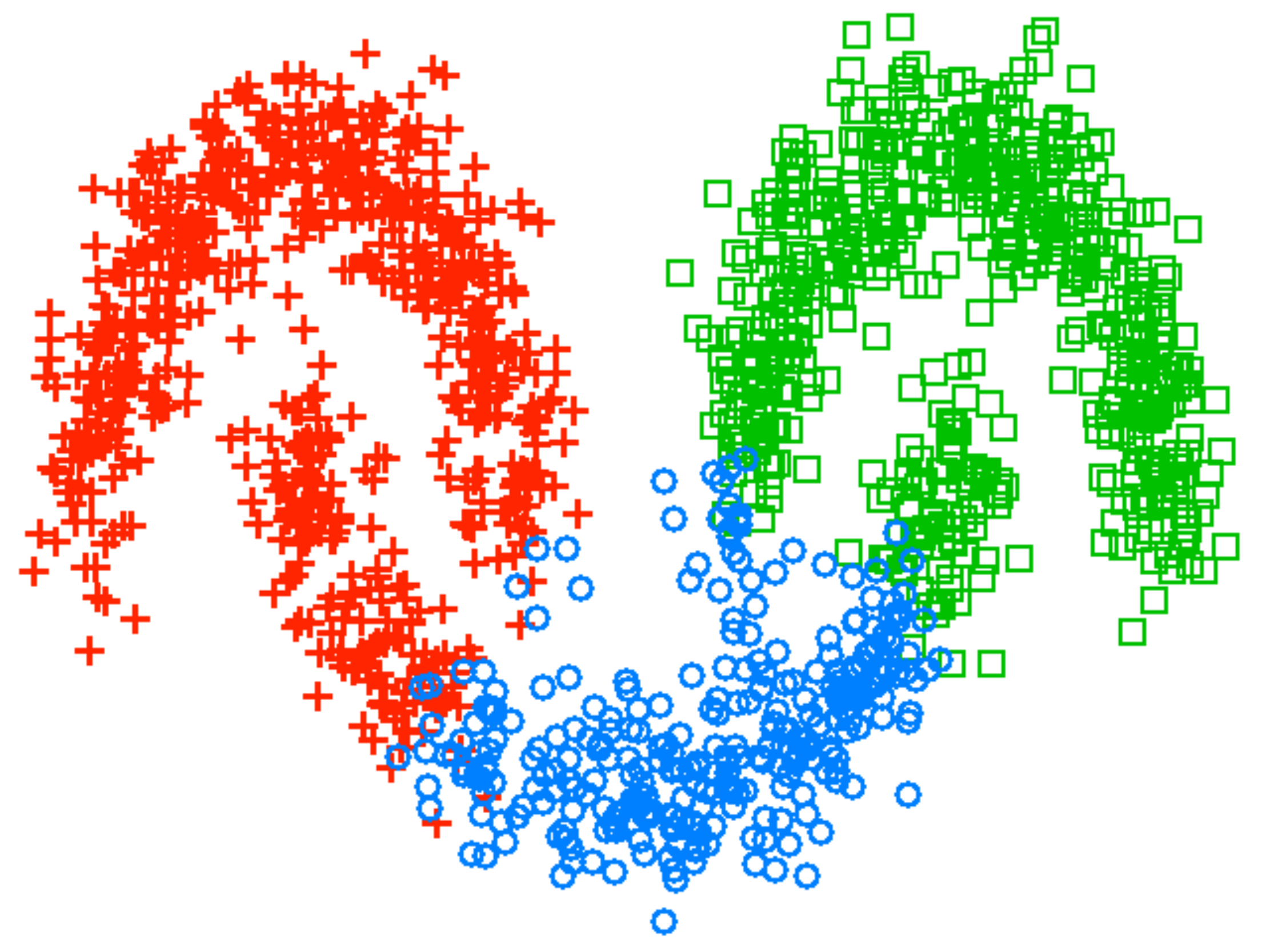}}
                \caption{$\epsilon=2.5$}
        \end{subfigure}
         \begin{subfigure}[b]{0.24\textwidth}
                \centering
                     \scalebox{0.4}{\includegraphics[clip,width=3.5in]{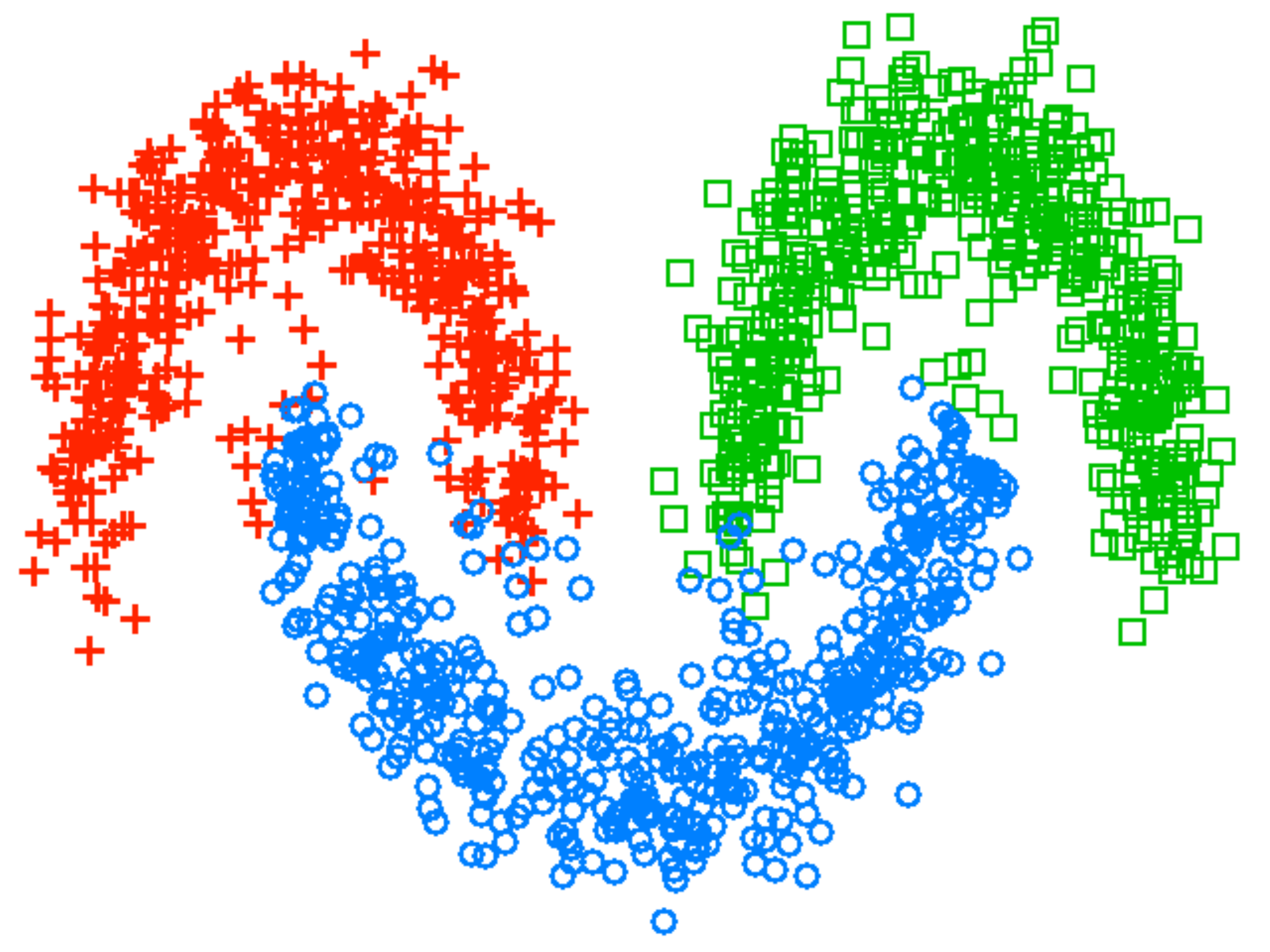}}
                \caption{$\epsilon=1$}
        \end{subfigure}
\caption{Three-moons segmentation. Left: $\epsilon = 2.5$ (81.8\% correct). Right: $\epsilon =1$ (97.1 \% correct).}
\label{fig:3moon_eps}
\end{figure}

The parameter $\epsilon$ determines a scale for the diffuse interface
and therefore has consequences in the minimization of the GL energy
functional, as discussed in Section~\ref{ssl_graphs}. Smaller values of
$\epsilon$ define a smaller length for the diffuse interface, and at the
same time, increasing the relative weight of the potential term with
respect to the smoothing term. Therefore, as the parameter $\epsilon$
decreases, sharp transitions are generated which in general constitute
more accurate classifications. Figure~\ref{fig:3moon_eps} compares the
performance for two different values of $\epsilon$. Note that the GL
results for large $\epsilon$ are roughly comparable to those given by a
standard spectral clustering approach~\cite{gc:flenner:percus}.

\subsection{Co-segmentation}
We tested our algorithms on the task of co-segmentation. In this task, two images with a similar topic are used. On one of the images, several regions are labeled. The image labeling task looks for a procedure to transfer the knowledge about regions, specified by the labeled segmentation, onto the unlabeled image. Thus, the limited knowledge about what defines a region is used to segment similar images without the need for further labelings.

\begin{figure}[!h]
         \begin{subfigure}[b]{0.24\textwidth}
                \centering
                     \scalebox{0.47}{\includegraphics[clip,width=3.25in,height=2.5in]{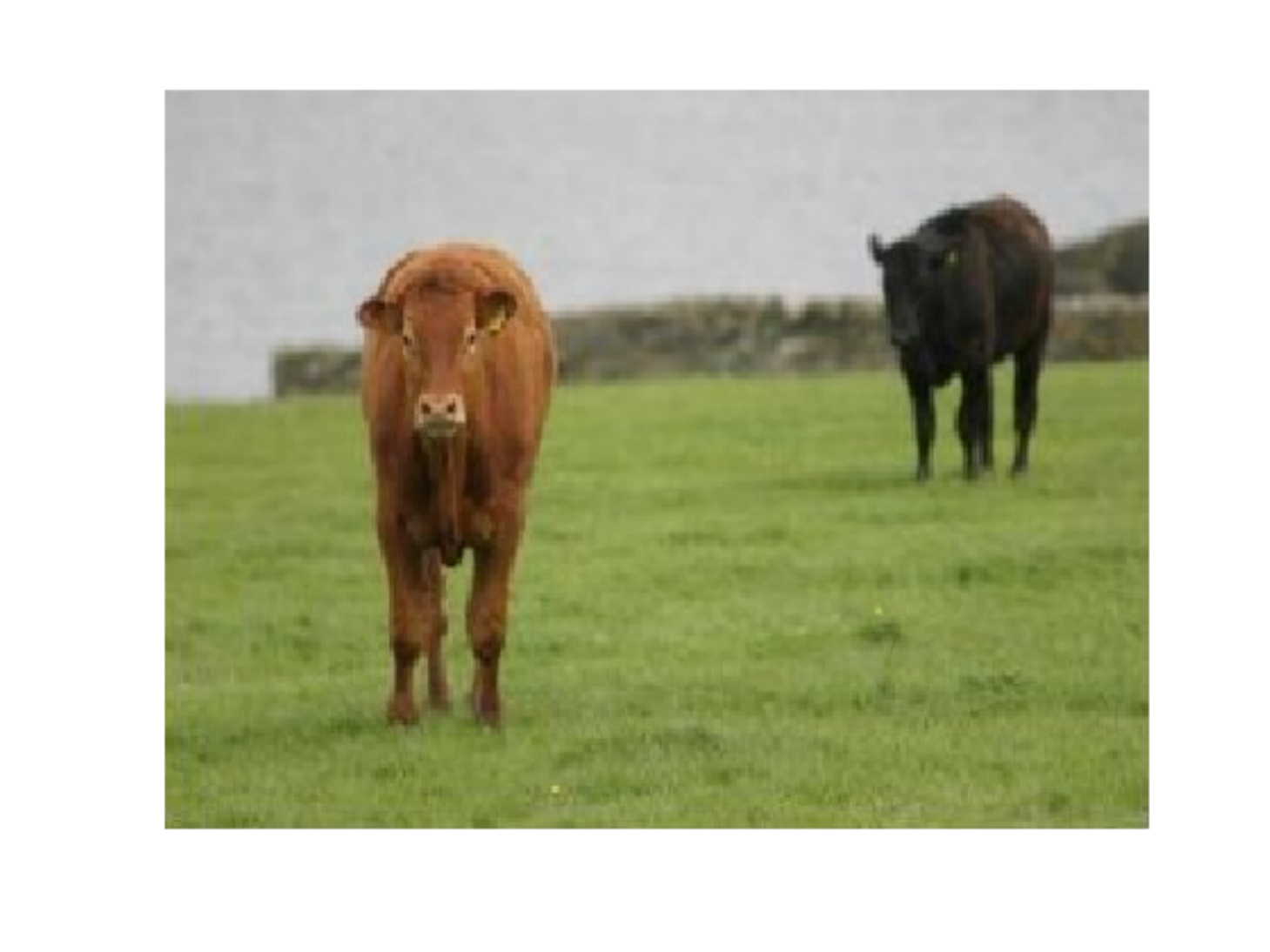}}
               \caption{Original Image}
                \label{fig:cows_}
        \end{subfigure}
         \begin{subfigure}[b]{0.24\textwidth}
                \centering
                     \scalebox{0.47} {\includegraphics[clip,width=3.25in,height=2.5in]{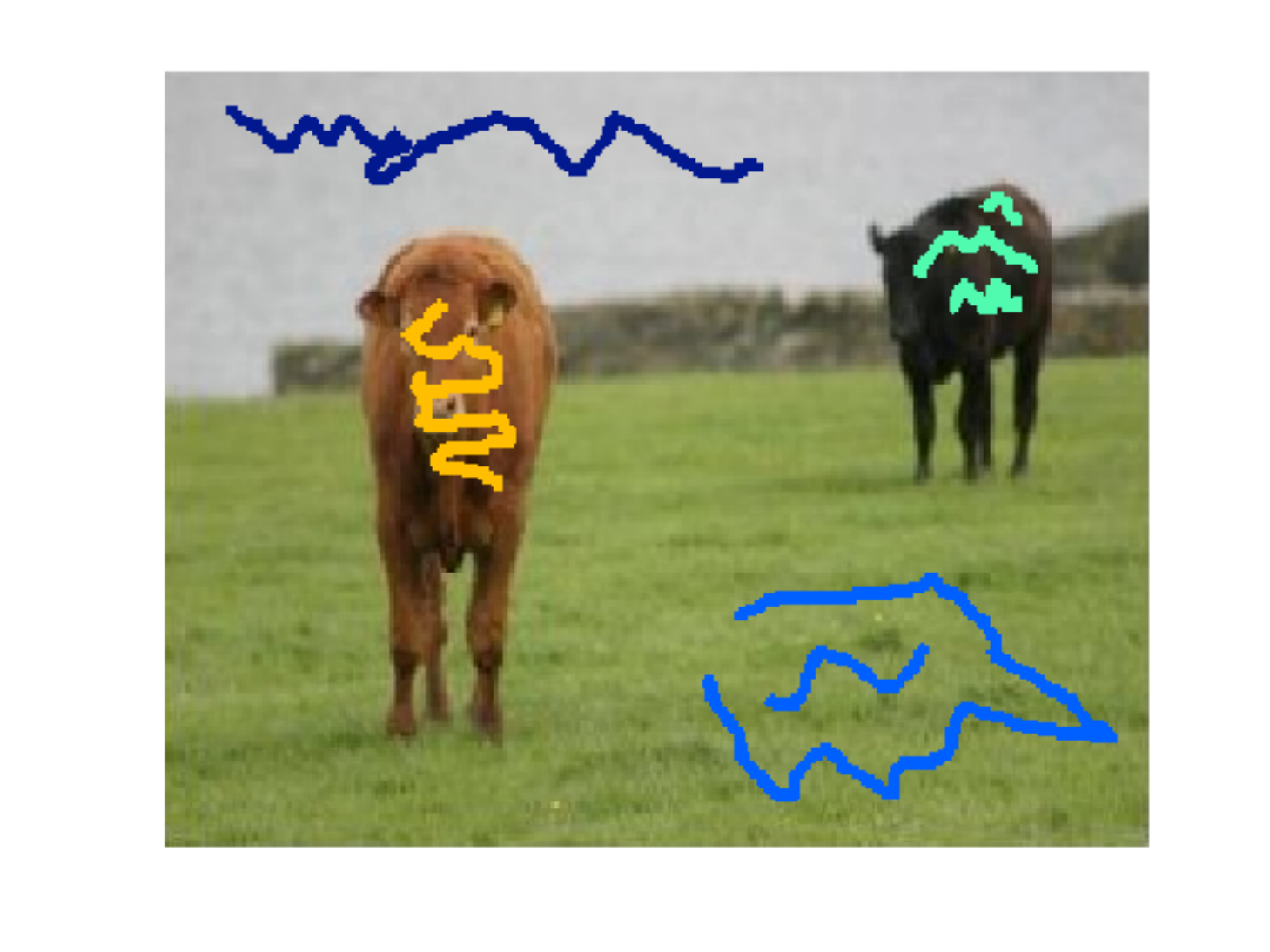}}
		\caption{Labeled Data}
                \label{fig:cows_in}
        \end{subfigure}
                
\caption{Labeled Color Image}
\label{fig:cow_labeled}
\end{figure}

\begin{figure}[!h]
        \begin{subfigure}[b]{0.5\textwidth}
               \centering
                 \scalebox{0.47} {\includegraphics[clip,width=3.25in]{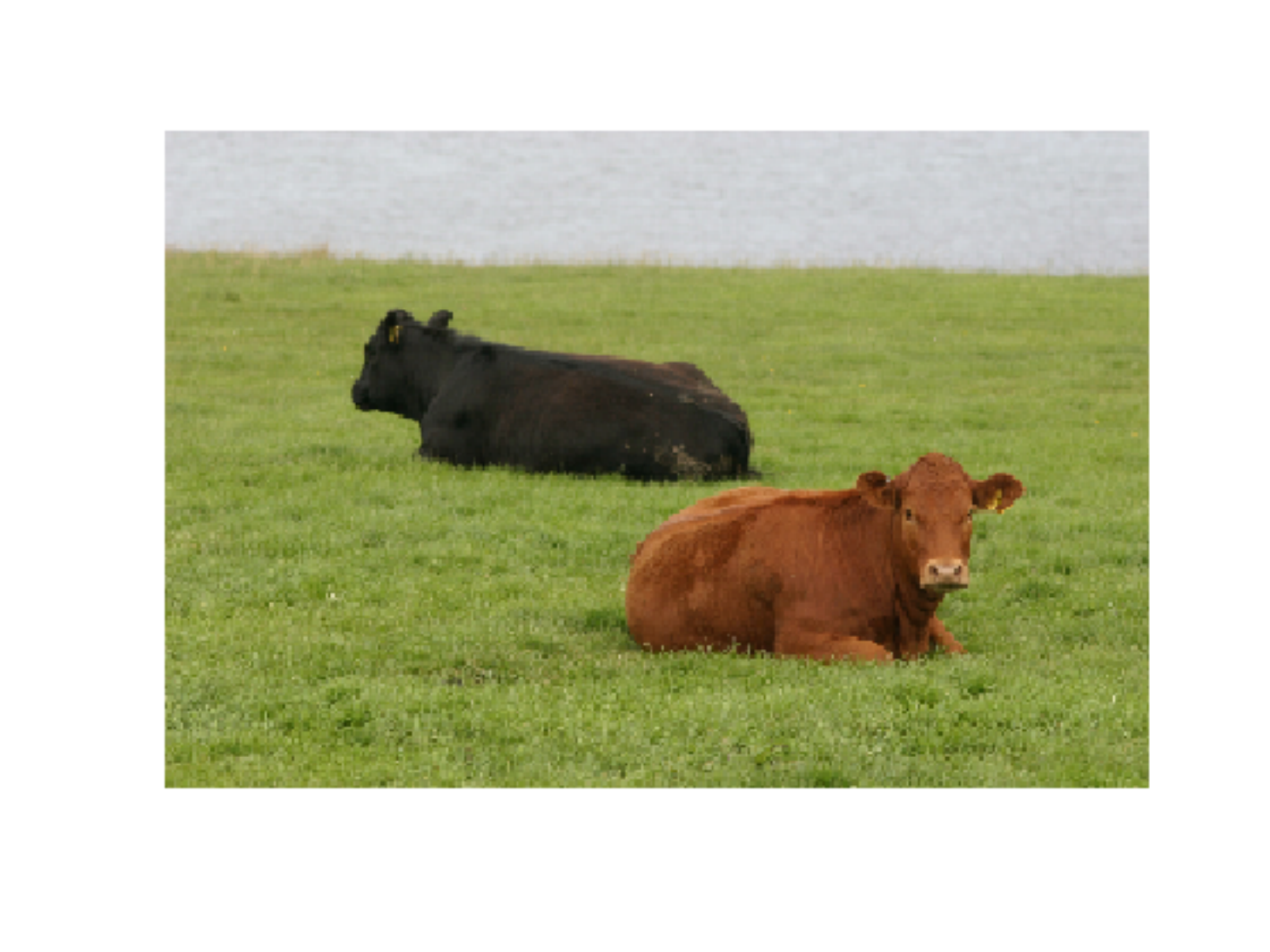}}
                \caption{Image to Segment}
                \label{fig:cows_to}
        \end{subfigure}
        
         \begin{subfigure}[b]{0.24\textwidth}
                \centering
                     \scalebox{0.47}{\includegraphics[clip,width=3.25in,height=2.5in]{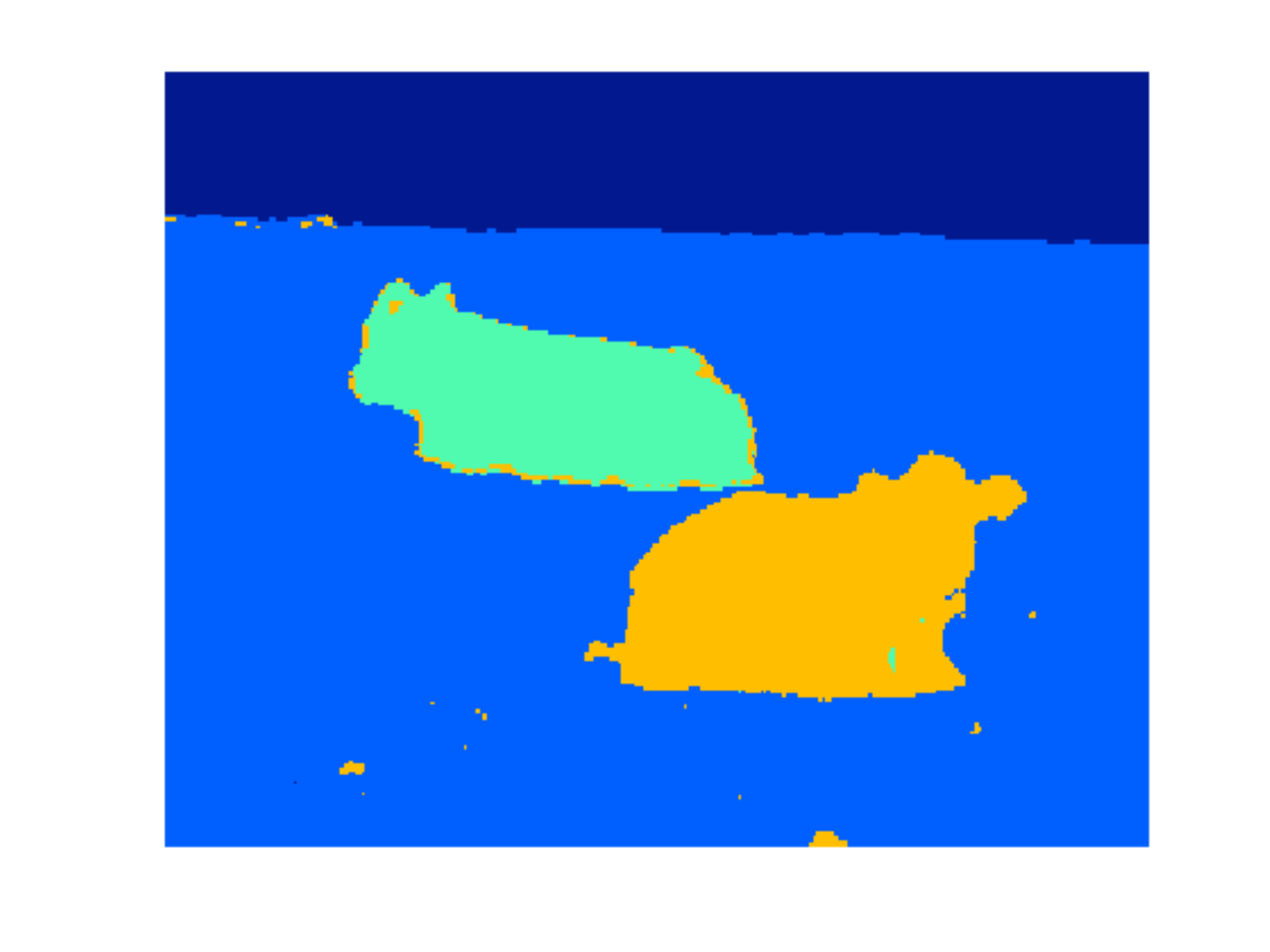}}
                \caption{Multiclass GL}
                \label{fig:cow_MGL}
        \end{subfigure}
         \begin{subfigure}[b]{0.24\textwidth}
                \centering
                     \scalebox{0.47} {\includegraphics[clip,width=3.25in,height=2.5in]{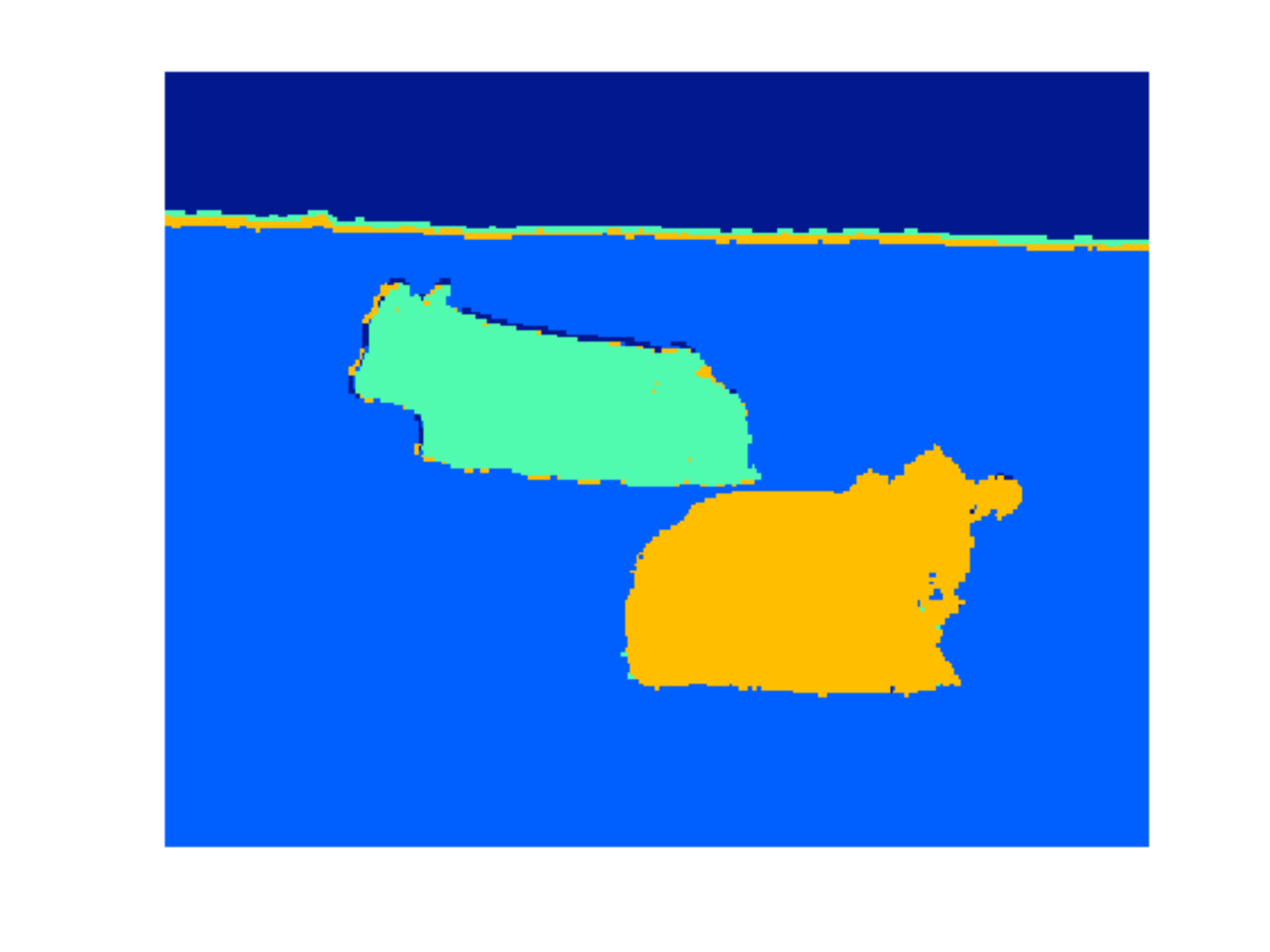}}
		\caption{Multiclass MBO}
                \label{fig:cow_MBO}
        \end{subfigure}        

\caption{Resulting Color Image Segmentation}
\label{fig:cow_result}
\end{figure}

On the color image of cows, shown in Figure~\ref{fig:cows_}, some parts of the sky, grass, black cow and red cow have been labeled, as shown in Figure~\ref{fig:cows_in}. This is a $319 \times 239$ color image.  The image to be segmented is a $319 \times 213$ color image shown in Figure~\ref{fig:cows_to}. The objective is to identify in this second image regions that are similar to the components in the labeled image.
 
 
To construct the weight matrix, we use feature vectors defined as the
set of intensity values in the neighborhood of a pixel. The neighborhood
is a patch of size $5 \times 5$. Red, green and blue channels are
appended, resulting in a feature vector of dimension 75. A Gaussian
similarity graph, as described in equation~(\ref{eq:gaussian}), is
constructed with $\sigma=22$ for both algorithms. Note that for both the
labeled and the unlabeled image, nodes that represent similar patches are
connected by high-weighted edges, independent of their position within
the image.  The transfer of information is then
enabled through the resulting graph, illustrating the nonlocal
characteristics of this unembedded graph-based method. 

The eigendecomposition of the Laplacian matrix is approximated using the
Nystr\"om method. This involves selecting 250 points randomly to
generate a submatrix, whose eigendecomposition is used in combination
with matrix completion techniques to generate the approximate
eigenvalues for the complete set. Details of the Nystr\"om method are
given
elsewhere~\cite{bertozzi:flenner,fowlkes:belongie:malik,fowlkes:belongie:chung}.
This approximation drastically reduces the computation time, as seen in Table~\ref{timing}.


The multiclass Ginzburg-Landau method used the following parameters: $200$
eigenvectors, $\epsilon=1$, $dt=0.005$, $\mu=50$ and
$\eta=10^{-7}$. 

The multiclass MBO method used the following parameters: $250$ eigenvectors, $dt=0.005$, $\mu=300$, $\eta=10^{-7}$.  

One of the results of each of our two methods (using the same fidelity set) is
depicted in Figure~\ref{fig:cow_result}. It can be seen that both
methods are able to transfer the identity of all the classes, with
slightly better results for mutliclass MBO.  Most of the mistakes made
correspond to identifying some borders of the red cow as part of the
black cow. Multiclass GL also has problems identifying parts of the grass.

\subsection{MNIST Data}

\begin{figure}[h]
\centering
  \scalebox{1.1}  {\includegraphics[trim=50 40 50 20,clip,width=2in,height=1in]{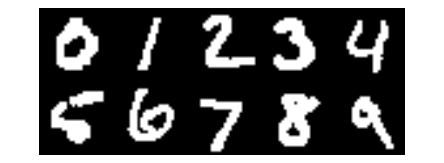}}
\caption{Examples of digits from the MNIST data base}
\label{fig:digits}
\end{figure}

The MNIST data set~\cite{lecun:cortes} is composed of $70,000$ $28
\times 28$ images of handwritten digits $0$ through $9$.
Examples of entries can be found in Figure \ref{fig:digits}. The task is to classify each of the images into the corresponding digit. The images include digits from $0$ to $9$; thus, this is a $10$ class segmentation problem.

To construct the weight matrix, we used $N = 8$ nearest neighbors with
local scaling based on the $8^{th}$ closest neighbor ($M = 8$). Note
that we perform no preprocessing, i.e. the graph is constructed using the $28 \times 28$ images. For the fidelity term, $250$ images per class ($2500$ images corresponding to $3.6 \%$ of the data) are chosen randomly.

The multiclass GL method used the following parameters: $300$
eigenvectors, $\epsilon=1$, $dt=0.15$, $\mu=50$ and
$\eta=10^{-7}$. The set of 70,000 images was segmented with an average accuracy (over $10$ runs) of $96.8$\% of the digits classified correctly in an average time of $153$~s. 

The multiclass MBO method used the following parameters: $300$
eigenvectors, $dt=0.15$, $\mu=50$, $\eta=10^{-7}$. The algorithm segmented an average of $96.91$\% of the digits correctly over $10$ runs in only $4$ iterations and $15.382$~s. We display the confusion matrix in Table \ref{confusion_MBO}. Note that most of the mistakes were in distinguishing digits $4$ and $9$, and digits $2$ and $7$. 

\begin{table*}[t]
\centering
\caption{Confusion Matrix for MNIST Data Segmentation: MBO Scheme}
\label{confusion_MBO}

\renewcommand{\arraystretch}{1.3}

\begin{tabular}{|c||c| c| c| c| c| c| c| c| c| c|}
\hline
Obtained/True & 0 & 1 & 2 & 3 & 4 & 5 & 6 & 7 & 8 & 9 \\
\hline
0 & 6844 & 20 & 41 & 3 & 3 & 15 & 21 & 1 & 20 & 17\\
\hline
1 & 5 & 7789 & 32 & 8 & 34 & 1 & 14 &  63 & 51 & 14\\
\hline
2 & 5 & 22 & 6731 & 42 & 2 & 4 & 1 & 23 & 19 & 8\\
\hline
3 & 0 & 3 & 20 & 6890 & 1 & 86 & 0 & 1 & 81 & 90\\
\hline
4 & 1 & 17 & 6 & 2 & 6625 & 3 & 7 & 12 & 28 & 67\\
\hline
5 & 9 & 0 & 3 & 70 & 0 & 6077 & 28 & 2 & 109 & 14\\
\hline
6 & 31 & 5 & 11 & 3 & 22 & 69 & 6800 & 0 & 29 & 5\\
\hline
7 & 2 & 16 & 117 & 44 & 12 & 9 & 0 & 7093 & 20 & 101\\
\hline
8 & 2 & 2 & 21 & 46 & 4 & 17 & 5 & 2 & 6398 & 22\\
\hline
9 & 4 & 3 &  8 & 33 & 121 & 32 & 0 & 96 & 70 & 6620\\
\hline
\end{tabular}
\end{table*}

Table~\ref{benchmark} compares our results with those from other methods
in the literature.  As with the three moon problem, some of these
are based on unsupervised methods but incorporate enough prior
information that they can fairly be compared with SSL methods.  The
methods of linear/nonlinear classifers, $k$-nearest neighbors, boosted
stumps, neural and convolutional nets and SVM are all supervised learning
approaches, taking 60,000 of the digits as a training set and 10,000
digits as a testing set~\cite{lecun:cortes}, in comparison to our SSL
approaches where we take
only $3.6\%$ of the points for the fidelity term.  Our algorithms are
nevertheless competitive with, and in most cases outperform, these
supervised methods.
Moreover, we perform no preprocessing or initial feature extraction on
the image data,
unlike most of the other methods we compare with (we exclude from
the comparison the methods that deskewed the image).
While there is a
computational price to be paid in forming the graph when data points
use all 784 pixels as features (see graph calculation time in
Table~\ref{timing}), this is a one-time operation that conceptually
simplifies our approach.
%

\subsection{COIL dataset}

We evaluated our performance on the benchmark COIL data
set~\cite{coil,chapelle:scholkopf:zien}.
This is a set of color $128 \times 128$  images of $100$ objects, taken at different angles. The red channel of each image was then downsampled to $16 \times 16$ pixels by averaging over blocks of $8 \times 8$ pixels. Then $24$ of the objects were randomly selected and then partitioned into six classes. Discarding $38$ images from each class leaves $250$ per class, giving a data set of $1500$ data points. 

To construct the weight matrix, we used $N = 4$ nearest neighbors with
local scaling based on the $4^{th}$ closest neighbor ($M = 4$). The
fidelity term was constructed by labeling $10$\% of the points, selected
at random.

For multiclass GL, the parameters were: 
$35$ eigenvectors, $\epsilon=1$, $dt=0.05$, $\mu=50$ and $\eta=10^{-7}$. This resulted in 91.2\% of the points classified correctly (average) in $0.035$~s.

For multiclass MBO, the parameters were: $50$ eigenvectors, $dt=0.2$,
$\mu=100$, $\eta=10^{-7}$. We obtained an accuracy of $91.46$\%, averaged over $10$ runs. The procedure took $6$ iterations and $0.03$~s. 

Comparative results reported in~\cite{subramanya} are
shown in Table~\ref{benchmark}.  These are all SSL methods (with the
exception of $k$-nearest neighbors which is supervised), using $10\%$
fidelity just as we do.  Our results are of comparable or greater accuracy.


\subsection{WebKB dataset}

Finally, we tested our methods on the task of text classification on the
WebKB data set~\cite{webKB}.  This is a collection of webpages
from Cornell, Texas, Washington and Wisconsin universities, as well as other miscellaneous pages from other universities. The webpages are to be divided into four classes: project, course, faculty and student.
The data set is preprocessed as described in~\cite{cardoso}.

To construct the weight matrix, we used $575$ nearest neighbors. Tfidf
term weighting~\cite{cardoso} is used to represent the website feature vectors. They were then normalized to unitary length. The weight matrix points are calculated using cosine similarity.

For the multiclass GL, the parameters were: $250$ eigenvectors,
$\epsilon=1$, $dt=1$, $\mu=50$ and $\eta=10^{-7}$.
The average accuracies obtained for fidelity sets of different sizes are
given in Table~\ref{webkbfidelity}.
The average computation time was $0.49$~s.

For the multiclass MBO, the parameters were: $250$ eigenvectors,
$dt=1$, $\mu=4$, $\eta=10^{-7}$.
The average accuracies obtained for fidelity sets of different sizes are
given in Table~\ref{webkbfidelity}.
The procedure took an average of $0.05$ s and $7$ iterations. 

We compare our results with those of several supervised learning methods
reported in~\cite{cardoso}, shown in Table~\ref{benchmark}.
For these methods, two-thirds of the data were used for training, and one
third for testing. Our SSL methods 
obtain higher accuracy, using only $20$\% fidelity (for
multiclass MBO). Note that a larger sample of points for the
fidelity term reduces the error in the results, as shown
in Table~\ref{webkbfidelity}.
Nevertheless, the accuracy is high even for the smallest fidelity sets.
Therefore, the methods appear quite adequate for the SSL setting
where only a few labeled data points are known beforehand.

\paragraph*{Multiclass GL and MBO}

All the results reported point out that both multiclass GL
and multiclass MBO perform well in terms of data segmentation accuracy. 
While the ability to tune multiclass GL can be an advantage,
multiclass MBO is simpler and, in our examples, displays even better
performance in terms of its greater accuracy and the fewer number of iterations required.
Note that even though multiclass GL leads to the minimization of a non-convex function, in practice the results are comparable with other convex TV-based graph methods such as~\cite{bresson:laurent:2013}. 
Exploring the underlying connections of the energy evolution of these methods and the energy landscape for the relaxed Cheeger cut minimization recently established in~\cite{bresson:laurent} are to be explored in future work.

\section{Conclusions}
\label{conclusions}

We have presented two graph-based algorithms for multiclass
classification of high-dimensional data. The two algorithms are based
on the diffuse interface model using the Ginzburg-Landau functional, and
the multiclass extension is obtained using the Gibbs simplex. The first algorithm minimizes the functional using gradient descent and a
convex-splitting scheme. The second algorithm executes a simple
scheme based on an adaptation of the classical numerical MBO
method. It uses fewer parameters than the first algorithm, and while
this may in some cases make it more restrictive, in our experiments it
was highly accurate and efficient.  

Testing the algorithms on synthetic data, image labeling and benchmark data sets
shows that the results are competitive with or better than some of the
most recent and best published algorithms in the literature. In addition,
our methods have several advantages. First, they are simple and efficient,
avoiding the need for intricate function minimizations or heavy
preprocessing of data. Second, a relatively small proportion of fidelity
points is needed for producing an accurate result. For most of our data
sets, we used at most $10$\% of the data points for the fidelity term;
for synthetic data and the two images, we used no more than $5$\%.
Furthermore, as long as the fidelity set contains samples of all classes in
the problem, a random initialization is enough to produce good
multiclass segmentation results. Finally, our methods do not use
one-vs-all or sequences of binary segmentations that are needed for
some other multiclass methods. We therefore avoid the bias and extra
processing that is often inherent in those methods.

Our algorithms can take advantage of the sparsity of the
neighborhood graphs generated by the local scaling procedure of
Zelnik-Manor and Perona~\cite{zelnik-manor}.  A further reason for the
strong practical performance of our methods is that the minimization
equations use only the graph Laplacian, and do not contain divergences
or any other first-order derivative terms. This allows us to use rapid
numerical methods. The Laplacian can easily be inverted by projecting
onto its eigenfunctions, and in practice, we only need to keep a small
number of these. Techniques such as the fast numerical
Rayleigh-Chebyshev method of Anderson~\cite{anderson} are very efficient
for finding the small subset of eigenvalues and eigenvectors needed.
In certain cases, we obtain
additional savings in processing times by approximating the
eigendecomposition of the Laplacian matrix through the Nystr\"om
method~\cite{bertozzi:flenner,fowlkes:belongie:malik,fowlkes:belongie:chung},
which is effective even for very large matrices: we need only compute
a small fraction of the weights in the graph, enabling the approximation
of the eigendecomposition of a fully connected weight matrix using
computations on much smaller matrices.

Thus, there is a significant
computational benefit in not having to calculate any first-order
differential operators.  In view of this, we have found that for general
graph problems, even though GL requires minimizing a non-convex
functional, the results are comparable in accuracy to convex
TV-based graph methods such as~\cite{bresson:laurent:2013}.
For MBO, the results are similarly accurate, with the further
advantage that the algorithm is very rapid.
We note that for other problems such as in image processing that
are suited to
a continuum treatment, convex methods and maxflow-type algorithms are in
many cases the best approach~\cite{chambolle:cremers:pock,yuan}.  It would be very interesting to try to extend our gradient-free numerical
approach to graph-based methods that directly use convex minimization, such as the method
described in \cite{yuan2}.

Finally, comparatively speaking, multiclass MBO performed better
than multiclass GL in terms of accuracy and convergence time for all of
the data sets we have studied.
Nevertheless, we anticipate that more intricate geometries could impair
its effectiveness. In those cases, multiclass GL might still perform
well, due to the additional control provided by tuning $\epsilon$ to
increase the thickness of the interfaces, producing smoother decision
functions.


\section*{Acknowledgments}

The authors are grateful to the reviewers for their comments and
suggestions, which helped to improve the quality and readability of the
manuscript. In addition, the authors would like to thank Chris Anderson
for providing the code for the Rayleigh-Chebyshev procedure of
\cite{anderson}. This work was supported by ONR grants N000141210838,
N000141210040, N0001413WX20136, AFOSR MURI grant FA9550-10-1-0569, NSF
grants DMS-1118971 and DMS-0914856, the DOE Office of Science's ASCR
program in Applied Mathematics, and the W. M. Keck Foundation. Ekaterina Merkurjev is also supported
by an NSF graduate fellowship.

\ifCLASSOPTIONcaptionsoff
  \newpage
\fi

\nocite{*}
\bibliographystyle{IEEEtranS}
\bibliography{IEEEabrv,multisegmentation}

\begin{IEEEbiography}[{\includegraphics[width=1in,height=1.25in,clip,keepaspectratio]{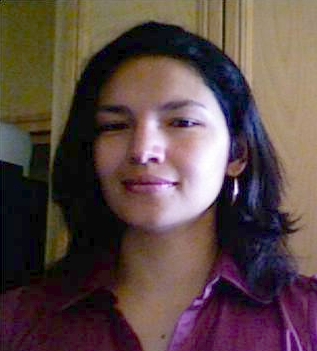}}]{Cristina Garcia-Cardona}
is a Postdoctoral Fellow at Claremont Graduate University. She obtained her Bachelor's degree in Electrical Engineering from Universidad de Los Andes in Colombia and her Master's degree in Emergent Computer Sciences from Universidad Central de Venezuela. Recently, she received her PhD in Computational Science from the Claremont Graduate University and San Diego State University joint program, working under the supervision of Prof. Allon Percus and Dr. Arjuna Flenner. Her research interests include energy minimization and graph algorithms.
\end{IEEEbiography}

\begin{IEEEbiography} [{\includegraphics[width=1in,height=1.25in,clip,keepaspectratio]{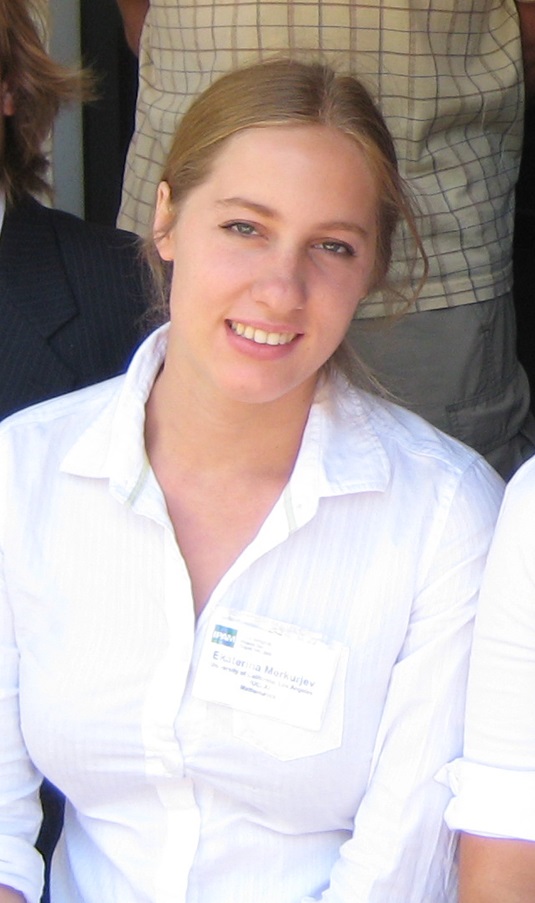}}]{Ekaterina Merkurjev}
is a fourth year graduate student at the UCLA
Department of Mathematics. She obtained her Bachelors and Masters
degrees in Applied Mathematics from UCLA in 2010. She is currently
working on a PhD under the supervision of Prof.\ Andrea Bertozzi.
Her research interests include image processing and segmentation. 
\end{IEEEbiography}

\begin{IEEEbiography}[{\includegraphics[width=1in,height=1.25in,clip,keepaspectratio]{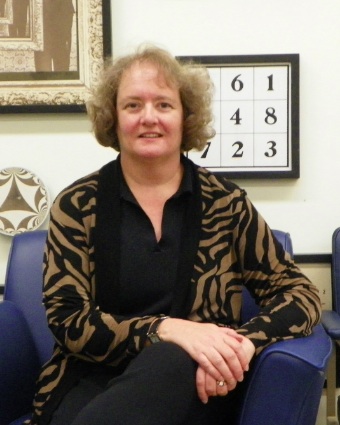}}]{Andrea L. Bertozzi}
received the BA, MA, and PhD degrees in mathematics from Princeton University, Princeton, NJ, in 1987, 1988, and 1991 respectively.
She was on the faculty of the University of Chicago, Chicago, IL, from 1991-1995 and Duke University, Durham, NC, from 1995-2004. During
1995-1996, she was the Maria Goeppert-Mayer Distinguished Scholar at Argonne National Laboratory. Since 2003, she has been with the University of California,
Los Angeles, as a Professor of Mathematics and currently serves as the Director of Applied Mathematics. In 2012 she was appointed the Betsy Wood Knapp Chair for Innovation and Creativity.
Her research interests include image inpainting, image segmentation, cooperative control of robotic vehicles, swarming, and fluid interfaces, and crime modeling.
Prof.\ Bertozzi is a Fellow of both the Society for Industrial and Applied Mathematics and the American Mathematical Society; she is a member of the American Physical Society.
She has served as a Plenary/Distinguished Lecturer for both SIAM and AMS and is an Associate Editor for the SIAM journals Multiscale Modelling and Simulation,
Mathematical Analysis. She also serves on the editorial board of Interfaces and Free Boundaries, Applied Mathematics
Research Express, Nonlinearity, Appl. Math. Lett., Math. Mod. Meth. Appl. Sci. (M3AS), J. Nonlinear Sci, J. Stat. Phys., Comm. Math. Sci., Nonlinear Anal. Real World Appl., and Adv. Diff. Eq. Her past honors include a Sloan Foundation Research Fellowship, the Presidential Career Award for Scientists and Engineers, and the SIAM Kovalevsky Prize in 2009.
\end{IEEEbiography}

\begin{IEEEbiography}[{\includegraphics[width=1in,height=1.25in,clip,keepaspectratio]{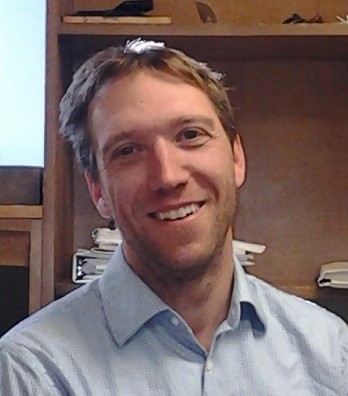}}]{Arjuna Flenner}
received his Ph.D. in Physics at the University of
Missouri-Columbia in 2004. His major
emphasis was mathematical physics. Arjuna Flenner's research interests
at the Naval Air Weapons Centre at China Lake include image
processing, machine learning, statistical pattern recognition, and
computer vision. In particular, he has investigated automated image
understanding algorithms for advanced naval capabilities.
His main research areas are nonlocal operators, geometric
diffusion, graph theory, non-parametric Bayesian analysis, and
a-contrario hypothesis testing methods. Arjuna Flenner was a US
Department of Energy GAANN Fellowship in 1997-2001, and currently is also
a visiting research professor at Claremont Graduate University.
\end{IEEEbiography}

\begin{IEEEbiography}[{\includegraphics[width=1in,height=1.25in,clip,keepaspectratio]{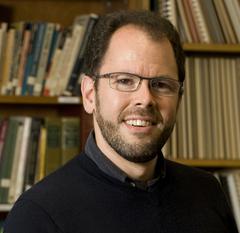}}]{Allon G.\ Percus}
received his BA in physics from Harvard in 1992 and his
PhD from the Universit\'{e} Paris-Sud, Orsay in 1997.  He was a member
of the scientific staff at Los Alamos National Laboratory in the
Division of Computer and Computational Sciences, and from 2003 to 2006
he served as Associate Director of the Institute for Pure and Applied
Mathematics at UCLA.  Since 2009, he has been Associate Professor of
Mathematics at Claremont Graduate University.  His research interests
combine discrete optimization, combinatorics and statistical physics,
exploiting physical models and techniques to study the performance of
algorithms on NP-hard problems.
\end{IEEEbiography}

\end{document}